\documentclass[11pt]{article}

\usepackage{amsmath, amssymb, amsthm}
\usepackage{graphicx}
\usepackage{float}
\usepackage[numbers]{natbib}
\usepackage{geometry}
\usepackage{setspace}
\usepackage{hyperref}
\hypersetup{
    hidelinks,
    pdfborder={0 0 0}
}
\usepackage{tikz}
\usepackage{pgfplots}
\pgfplotsset{compat=1.18}
\usetikzlibrary{calc}
\usetikzlibrary{shapes}
\usetikzlibrary{arrows.meta}
\usetikzlibrary{positioning}
\usetikzlibrary{decorations.pathmorphing}

\usepackage{algorithm} 
\usepackage{algpseudocode}
\usepackage{booktabs}
\usepackage{subcaption}

\algrenewcommand\algorithmiccomment[1]{\hfill{\scriptsize\(\triangleright\) #1}}
\theoremstyle{plain}
\newtheorem{theorem}{Theorem}[section]
\newtheorem{proposition}[theorem]{Proposition}

\theoremstyle{definition}

\theoremstyle{remark}
\newtheorem{remark}[theorem]{Remark}
\newcommand{\keywords}[1]{%
  \vspace{1em}
  \noindent\textbf{Keywords: } \small{#1}
}

\title{The Blueprints of Intelligence\\[0.3em]
       \large A Functional–Topological Foundation for Perception and Representation}

\author{
Eduardo Di Santi.  
\small{University of Colorado Boulder} \\
}
\date{\today}

\begin{document}
\maketitle

\begin{abstract}
Real-world phenomena do not generate arbitrary variability: 
their signals concentrate on compact, low-variability subsets of functional space, enabling rapid generalization from few examples. 
We formalize this principle through a deterministic functional--topological framework in which the set of valid realizations produced by a physical process forms a compact subset of a Banach space, endowed with stable invariants, a finite Hausdorff radius, and an induced continuous perceptual functional. 
This geometry provides structural constraints on variability, conditions for identifiability, and supports generalization from sparse evidence.

We develop this framework in full generality and then examine its empirical relevance across 
five real-world domains. In electromechanical (railway point machines), electrochemical (battery 
discharge), physiological (ECG), atmospheric (solar irradiance), and geophysical (tidal cycles) 
systems, and, where available, in corresponding deterministic simulators, we show that the empirical 
Hausdorff radius of the perceptual manifold saturates after surprisingly few samples, indicating 
that the admissible signal families occupy compact, low-variability regions of function space.

Across all these settings, geometric saturation emerges rapidly and consistently, suggesting that compact perceptual manifolds provide a powerful organizing principle for both physical processes and learned representations. 
Together, these results support deterministic functional topology as a promising framework for understanding perception and representation.
\end{abstract}


\section{Introduction}

Understanding why biological learners can generalize from only a handful of
observations requires examining the structure of the signals produced by the
physical world.
Real phenomena do not generate arbitrary variability: their signals concentrate
around low-dimensional, compact subsets of functional space shaped by the
underlying physics.
This geometric structure, rather than data quantity alone, helps explain why
perception can be rapid, robust, and sample-efficient.

This framework is not merely a mathematical account of real-world signals;
it articulates a structural basis for perception and representation.
If perception consists in identifying compact manifolds of admissible realizations,
then intelligent systems can be understood, at least in their perceptual layer,
as geometric observers of the world rather than solely as statistical predictors
trained on arbitrary datasets.
This perspective offers an alternative way to interpret learning systems:
as observers whose representations are strongly constrained by the geometry 
imposed by the physical world.

In deterministic systems, repeated measurements do not fill an unbounded space 
of possibilities; instead, they concentrate around a well-defined, 
low-variability structure in a functional space. 
This structure is inherently topological: the set of valid realizations 
generated by a physical system forms a compact subset of a Banach space, 
equipped with stable invariants, a finite Hausdorff radius, and a continuous 
functional that maps observations to compatibility scores. 
These properties impose strict limits on how much variability the world can 
exhibit and, consequently, on how much information an intelligent system must 
acquire to identify and distinguish real phenomena.

This viewpoint suggests that perception and representation can be studied as
problems of functional geometry, rather than solely as statistical approximation.
A perceptual category is not an arbitrary collection of samples, but a compact 
functional manifold with predictable boundaries and internal continuity. 
The ability to generalize from few examples arises naturally from this 
compactness: once the Hausdorff radius of the manifold has been explored, 
additional observations no longer expand the domain of valid realizations.

Throughout this work, $\mathcal{M}$ denotes the set of realized signals 
generated by the physical system—the observed perceptual manifold—rather than 
the full sensory space. 
When the governing equations are unknown, both the manifold structure and its 
radius must be inferred directly from the stream of observations, leading 
naturally to a self-supervised formulation. 
This view is consistent with contemporary approaches to autonomous intelligence 
\cite{lecun2022path}, which emphasize that learning arises from discovering the 
set of admissible representations produced by the world rather than from 
external supervision.

This paper develops a deterministic functional--topological framework for
perception and representation, and examines its consequences across several
real-world domains.
We show that the same geometric principles hold across distinct physical 
domains, including electromechanical, electrochemical, and physiological 
systems. 
In all cases, the signals generated by deterministic processes exhibit 
compactness, continuity, and stable invariants that allow their perceptual 
manifolds to be characterized and their boundaries to be estimated.

Our contributions are threefold:
\begin{enumerate}
  \item \textbf{Functional–topological framework.}
  We introduce a topological framework in which real-world perceptual sets
  are modeled as compact subsets of $C^0([0,T])$ with finite Hausdorff
  radius and stable invariants, and we show how deterministic physical
  processes induce continuous perceptual functionals on these manifolds%
  \cite{rudin1991functional, royden2010real}.
  \item \textbf{Geometric saturation in physical and biological systems.}
  We provide empirical evidence that deterministic functional topology accurately
   describes several real-world signal families across electromechanical (railway point machines),
   electrochemical (battery discharge), physiological (ECG), atmospheric (solar irradiance),
   and geophysical (tidal cycles) systems, together with their deterministic simulators. 
   In all cases, the empirical Hausdorff radius of the perceptual manifold saturates after surprisingly few samples,
   indicating that admissible signals occupy compact, low-variability regions of function space.
  \item \textbf{Extension to perceptual benchmarks and algorithmic implications.}
  Using the same pipeline, we observe analogous geometric saturation on
  standard perceptual benchmarks (MNIST and Spoken MNIST) when classes are
  treated as compact manifolds and inference is based on topological
  proximity. This cross-domain consistency suggests that compact perceptual
  manifolds provide a useful organizing principle for both physical processes
  and learned representations, and it motivates algorithms that explicitly
  exploit early saturation of the perceptual radius (e.g., in similarity
  search and matrix-profile–based analysis of deterministic time series).
\end{enumerate}

To illustrate these ideas, Figure~\ref{fig:perceptual_manifold} shows a schematic representation of a 
perceptual manifold embedded in a closed ball of $C^0([0,T])$, together with 
examples drawn from the five domains studied in this work. Although the 
physical processes differ, their signal manifolds share the same compact, 
low-variability geometry.

\begin{figure}[h!]
\centering
\begin{tikzpicture}[x=1cm,y=1cm]


\tikzset{mini plot/.style={draw=black, very thin, minimum width=3cm, minimum height=1cm}}

\begin{scope}[shift={(-4,1.5)}]
  \node[mini plot, anchor=west] (pmframe) at (0,0) {};
  \begin{scope}[shift={(0,-0.5)}]
    \clip (0,0) rectangle (3,1);
    \draw[very thick]
      (0.1,0.1) -- (0.6,0.1)
      -- (0.8,0.9)
      -- (1.1,0.6)
      -- (2.4,0.6)
      -- (2.8,0.1);
  \end{scope}
  \node[anchor=east] at (-0.2,0) {PM};
\end{scope}

\begin{scope}[shift={(-4,0)}]
  \node[mini plot, anchor=west] (batframe) at (0,0) {};
  \begin{scope}[shift={(0,-0.5)}]
    \clip (0,0) rectangle (3,1);
    \draw[very thick]
      (0.1,0.8) .. controls (1.2,0.75) and (2.0,0.4) .. (2.8,0.25);
  \end{scope}
  \node[anchor=east] at (-0.2,0) {Battery};
\end{scope}

\begin{scope}[shift={(-4,-1.5)}]
  \node[mini plot, anchor=west] (ecgframe) at (0,0) {};
  \begin{scope}[shift={(0,-0.5)}]
    \clip (0,0) rectangle (3,1);
    \draw[very thick]
      (0.1,0.4) -- (0.5,0.4)
      .. controls (0.6,0.5) .. (0.7,0.4)
      -- (0.9,0.4) -- (1.0,0.2)
      -- (1.05,0.9)
      -- (1.1,0.3)
      -- (1.4,0.4)
      .. controls (1.6,0.6) and (1.9,0.5) .. (2.1,0.4)
      -- (2.8,0.4);
  \end{scope}
  \node[anchor=east] at (-0.2,0) {ECG};
\end{scope}


\begin{scope}[shift={(1.3,0)}]

\shade[ball color=gray!10, opacity=1] (0,0) circle (2.2);

\draw[blue!80, line width=2pt]
  (0.9,0.1)
    .. controls (0.7,0.9) and (0.0,1.1) ..
  (-0.7,0.7)
    .. controls (-1.3,0.2) and (-1.1,-0.7) ..
  (-0.2,-0.9)
    .. controls (0.6,-0.9) and (1.2,-0.4) ..
  (0.9,0.1);  

\fill (0,0) circle (1.3pt);
\node[anchor=north east] at (0,0) {$x_0$};

\draw[very thick] (0,0) -- (0.9,0.2);
\node[anchor=south west] at (0.3,0.1) {$r$};

\node[blue!80] at (0.2,1.2) {$\mathcal{M}$};

\node[font=\scriptsize] at (0,-2.7) {closed ball in $C^0([0,T])$};

\end{scope}


\draw[->, thin] (-1, 1.5) -- (0.6,0.8);    
\draw[->, thin] (-1, 0.0)  -- (0.2,0.0);    
\draw[->, thin] (-1,-1.5) -- (0.4,-0.6);  

\end{tikzpicture}

\caption{\small
Schematic illustration of compact perceptual manifolds in $C^0([0,T])$.
Representative physical systems---railway point machines, battery discharge,
and ECG waveforms---generate signals that concentrate on compact,
low-variability manifolds. Other deterministic domains studied in this work,
including solar irradiance profiles and tidal cycles, follow the same geometric
principle: each domain has its own manifold, but all exhibit compact,
bounded-variability structure.
}
\label{fig:perceptual_manifold}
\end{figure}
The current paradigm in artificial intelligence relies heavily on scaling laws and
large datasets. However, biological intelligence often exhibits remarkable
sample efficiency. We hypothesize that part of this efficiency may arise from
geometric saturation in the underlying perceptual space. By applying the same
pipeline to standard benchmarks such as MNIST~\cite{lecun1998mnist} and Spoken
MNIST~\cite{khacef2020writtenspoken}, we observe that class-specific examples
also concentrate rapidly in compact regions of representation space. These
results suggest that topological representations may reduce the apparent data
requirements of learning systems when the underlying signal family possesses
compact structure.

\section{Related Work}
\label{sec:related}

Joint Embedding Predictive Architectures (JEPA) show that useful world representations can be learned by predicting latent embeddings instead of raw data. 
Recent vision--language instantiations such as VL-JEPA predict continuous target text embeddings from visual inputs and textual queries, achieving competitive or superior performance to classical token-generative VLMs under comparable training conditions~\cite{chen2025vljepa}. These approaches, however, do not provide a formal characterization of the geometry of the underlying signal manifolds that makes such prediction feasible, which is the focus of the present work.

\subsection{Industrial Assets: Railway Point Machines}

Railway point machines constitute a representative class of industrial electromechanical systems,
characterized by high-dimensional operational signals and strong physical constraints.
Recent work has demonstrated the effectiveness of deep learning--based approaches for scalable
diagnosis and predictive maintenance of point machines using large-scale operational data \cite{PM}.
However, these approaches primarily focus on classification and prognostics performance,
rather than on the intrinsic geometric structure of the space of physically admissible signals,
which is the focus of the present work.

\subsection{Physiological Signals: Electrocardiograms}

Electrocardiogram (ECG) signals have been extensively studied for arrhythmia detection,
morphological analysis, and clinical decision support using both classical signal processing
and machine learning techniques.
Despite the maturity of this literature, most existing approaches emphasize classification accuracy,
feature extraction, or diagnostic performance, rather than investigating the global geometric
structure and saturation properties of the space of admissible ECG waveforms.

\subsection{Electrochemical Systems: Battery Discharge Signals}

Battery discharge and degradation trajectories have been widely analyzed in the context of
state-of-health estimation, remaining useful life prediction, and aging analysis,
often using empirical electrochemical models or data-driven methods.
While these works successfully characterize temporal evolution and degradation trends,
they typically do not address whether the set of physically admissible discharge profiles
forms a compact geometric manifold with intrinsic saturation properties.

\section{Deterministic Systems and Perceptual Structure}

\subsection{Deterministic signal generation}

Let a physical system produce outputs $x(t)$ through a deterministic mapping
\[
x = f(s,\theta),
\]
where $s$ represents the internal physical state of the system and $\theta$ 
represents external conditions. Each realized signal $x(t)$ belongs to the 
Banach space $C^0([0,T])$ endowed with the supremum norm. Small perturbations,
manufacturing tolerances, and environmental variations are assumed bounded and 
preserve continuity.

We define the \emph{perceptual set} produced by the system:
\[
\mathcal{M} = \{\, f(s,\theta) : s\in\mathcal{S},\;\theta \in \Theta \,\}
\subset C^0([0,T]).
\]

\subsection{Compactness of the perceptual set}

A central premise of this work is that deterministic physical systems generate
signals that occupy a compact region of function space. This property is what
enables generalization from sparse observations.

\begin{theorem}[Compactness of Deterministic Signals]
\label{thm:compactness}
If the family $\{f(s,\theta)\}$ is uniformly bounded and equicontinuous on 
$[0,T]$, then the perceptual set $\mathcal{M}$ is compact in 
$C^0([0,T])$.
\end{theorem}

\noindent Proof: See Appendix~A.

This result follows directly from the Arzelà–Ascoli theorem
\cite{royden2010real, rudin1991functional}. Compactness implies that the system 
cannot produce arbitrary variability: all realized signals lie inside a bounded,
closed, finite-variability geometric region in function space.

\subsection{Closed-ball structure and intrinsic invariants}

Compactness implies the existence of a center $x_0 \in \mathcal{M}$ 
and a finite radius $r$ such that:
\[
\mathcal{M} \subset B_\infty(x_0,r)
= \{\,x \in C^0 : \|x-x_0\|_\infty \le r \,\}.
\]

\begin{proposition}[Finiteness of the Perceptual Radius]
\label{prop:radius}
If $\mathcal{M}$ is compact, then the perceptual radius
\[
r := \sup_{x\in\mathcal{M}} \|x-x_0\|_\infty
\]
is finite.
\end{proposition}

\noindent Proof: See Appendix~A.

This radius represents the intrinsic extent of the phenomenon’s variability.
Within this ball, the system exhibits consistent geometric invariants: peaks,
plateaus, slopes, impact transients, or physiological wave morphology—features
that remain stable across realizations and thus serve as natural topological
identifiers of the underlying physical process.

\section{Perceptual Functions and the Universal Approximation Principle}

A perceptual process corresponds to mapping an observed signal to a numerical 
score, compatibility measure, or classification output. We formalize this as a 
continuous functional
\[
\Phi : \mathcal{M} \to \mathbb{R}
\]
defined on the compact perceptual manifold $\mathcal{M}$.

\subsection{Continuity of perceptual functionals}

Deterministic physical processes induce continuous variation of observations with
respect to changes in state or conditions. Thus, perceptual mappings that depend 
on physical structure (e.g.\ peak timing, amplitude, plateau stability) are 
naturally continuous in the supremum norm.

\begin{proposition}[Uniform Continuity on the Perceptual Manifold]
\label{prop:uniform}
If $\Phi$ is continuous on the compact set $\mathcal{M}$, then $\Phi$ is 
uniformly continuous on $\mathcal{M}$.
\end{proposition}

\noindent Proof: Heine–Cantor; see Appendix~A.

\subsection{Universal approximation of perceptual mappings}

A key implication of compactness and continuity is that perceptual functions are
universally approximable.

\begin{theorem}[Universal Approximation on a Compact Perceptual Manifold]
\label{thm:uat}
Let $\mathcal{M} \subset C^0([0,T])$ be compact and let
$\Phi : \mathcal{M} \to \mathbb{R}$ be continuous.
Then for every $\varepsilon > 0$ there exists a finite-dimensional embedding
$\pi_N : \mathcal{M} \to \mathbb{R}^N$ and a universal approximator
$N_\varepsilon : \mathbb{R}^N \to \mathbb{R}$ such that
\[
\sup_{x \in \mathcal{M}} |\Phi(x) - N_\varepsilon(\pi_N(x))| < \varepsilon.
\]
\end{theorem}

\noindent Proof: See Appendix~A.  
Follows from the Universal Approximation Theorem 
\cite{cybenko1989approximation, hornik1991approximation} and compactness of 
$\mathcal{M}$.

This shows that learnability arises from the compact geometry of the perceptual manifold: 
once the domain of admissible signals is compact and perceptual mappings are continuous, 
finite models suffice to approximate perception arbitrarily well.

\section{Empirical Radius and Knowledge Boundaries}

\subsection{The perceptual radius}

Using the Hausdorff metric $d_H$ \cite{edgar2008measure}, the perceptual 
radius is defined as:
\[
r = \sup_{x\in\mathcal{M}} d_H(\{x\}, \{x_0\})
= \sup_{x\in\mathcal{M}} \|x - x_0\|_\infty.
\]

The finiteness of $r$ follows directly from 
Proposition~\ref{prop:radius}.

\subsection{Monte Carlo estimation of the radius}

Sampling the physical system under varied $(s,\theta)$ provides empirical 
approximations of the supremum.

\begin{theorem}[Consistency of Monte Carlo Radius Estimation]
\label{thm:mc}
Let $(s_i,\theta_i)$ be samples whose support is dense in $\mathcal{S}\times\Theta$.
Define the estimator
\[
\hat{r}_n = \max_{1\le i \le n} \|f(s_i,\theta_i) - x_0\|_\infty.
\]
Then $\hat{r}_n \to r$ almost surely as $n\to\infty$.
\end{theorem}

\noindent Proof: See Appendix~A.

This provides a physical method for determining when the perceptual manifold 
has been fully explored.

\subsection{Identification as distance minimization}

Finally, classification or recognition reduces to computing the distance from an
observation to the perceptual manifold.

\begin{proposition}[Identification Criterion]
\label{prop:identification}
An observed signal $x$ is recognized as belonging to the phenomenon if and only if
\[
d_H(\{x\}, \mathcal{M}) < \varepsilon,
\]
for some tolerance $\varepsilon$ determined by the system's resolution.

\end{proposition}

\noindent Proof: See Appendix~A.

Thus, recognition is equivalent to minimum-distance classification in a compact 
functional space.

\subsection{Self-Supervised Emergence of the Perceptual Radius}

The perceptual radius $r$ plays a central role in determining the boundary of 
knowledge for a deterministic physical process. When the governing equations 
of the system are known, $r$ can be computed directly from the functional 
model:
\[
r = \sup_{s,\theta} \| f(s,\theta) - x_0 \|_\infty.
\]
However, in many real-world domains—electrochemical, physiological, or 
mechanical—the physical equations are partially known, high-dimensional, or 
altogether unavailable. In such cases, the observer must infer the perceptual 
structure directly from the observed signals.

\begin{proposition}[Self-Supervised Radius Identification]
\label{prop:selfsupervised}
Let $(x_i)_{i=1}^n$ be a sequence of realizations sampled from the physical 
process, and define the empirical radius
\[
\hat{r}_n = \max_{1 \le i \le n} \| x_i - x_0 \|_\infty.
\]
If sampling becomes dense in the underlying state–condition space, then 
$\hat{r}_n \to r$ almost surely. 
Thus, even without knowledge of the governing equations, the perceptual radius 
is recovered purely from observation.
\end{proposition}

\noindent Proof: Follows directly from Theorem~\ref{thm:mc} and compactness of 
$\mathcal{M}$; see Appendix~A.

\begin{remark}[Self-Supervised Perception]
The convergence of $\hat{r}_n$ implies that the observer operates in a 
self-supervised regime: the perceptual manifold $\mathcal{M}$ and its radius 
are discovered directly from the stream of observations, without labels, 
external supervision, or prior knowledge of the dynamics. 
As variability is exhausted, the manifold stabilizes and the radius saturates.
\end{remark}

This property is evident across the five domains studied here. For railway 
point machines, where the physical model is partially known, the theoretical 
bounds and empirical estimates agree. For battery discharge curves and ECG 
signals, where the underlying equations are largely inaccessible, the empirical 
radius exhibits natural convergence, revealing the compact structure of the 
perceptual manifold directly from data.

\section{Methods}
\label{sec:methods}
Our experimental evaluation follows a unified pipeline applied identically 
across the five physical domains studied in this work. The goal is to 
estimate the geometry of the perceptual manifold---its compactness, invariants, 
and empirical radius---from real-world signals without relying on 
domain-specific modeling.

\subsection{Monte Carlo radius estimation}

To quantify how the perceptual radius evolves as sampling becomes dense, we 
estimate the empirical radius
\[
\hat r_n = \max_{1 \le i \le n} \|x_i - x_0\|_\infty
\]
over randomly drawn subsets of increasing size $n$.  
This procedure provides a nonparametric Monte Carlo (MC) estimator of the 
Hausdorff radius of the perceptual manifold.

Because the perceptual set $\mathcal{M}$ is compact, $\hat r_n$ is a 
monotonically non-decreasing sequence bounded above by the true radius $r$.  
Thus, as sampling becomes dense in the state--condition space, $\hat r_n$ 
converges to $r$ almost surely (Theorem~\ref{thm:mc}).  
The rate and shape of this convergence offer a practical diagnostic for 
manifold completeness: rapid initial growth reflects the discovery of previously
unseen variability, while the stabilization of $\hat r_n$ indicates that all 
extremal behaviours of the phenomenon have been observed.

Operationally, we compute $\hat r_n$ by repeatedly drawing random subsets of 
size $n \in \{10, 20, 50, \dots\}$, embedding signals in $\mathbb{R}^N$, 
and evaluating their distances to the reference signal $x_0$ under cosine geometry.
No labels, models, or physical assumptions are required; the estimator depends 
solely on the observed signals and therefore reflects the \emph{self-supervised
emergence} of the perceptual structure from data.

This MC-based procedure is applied identically to real datasets and to 
synthetic signals generated by the simulator, enabling direct comparison of 
radius saturation and geometric compactness under the same preprocessing and 
metric geometry.

\subsection{Preprocessing and functional normalization}

All signals are resampled onto a uniform temporal grid in $[0,T]$, detrended 
when necessary, and normalized to unit amplitude to ensure compatibility with 
the $C^0$ topology and the supremum norm. No temporal warping, smoothing, or 
feature extraction is applied.

\subsection{Distance metric and Hausdorff evaluation}

The theoretical framework is formulated in the Banach space $C^0([0,T])$ with 
the supremum norm $\|\cdot\|_\infty$, which induces the Hausdorff metric on 
compact subsets. 
Accordingly, the perceptual radius is defined as
\[
r = \sup_{x \in \mathcal{M}} \|x - x_0\|_\infty,
\]
and its empirical estimator is
\[
\hat{r}_n = \max_{1 \le i \le n} \|x_i - x_0\|_\infty.
\]

In practice, signals are discretized into vectors in $\mathbb{R}^N$ and 
stored in a vector index for efficient nearest-neighbor queries. 
On this finite-dimensional space, all norms are equivalent; therefore 
$\|\cdot\|_\infty$, $\|\cdot\|_2$, and cosine distance induce the same 
topology and the same notions of compactness and convergence. 
For the implementation, we use cosine distance on $\ell_2$-normalized vectors 
for efficient search, while retaining the $\|\cdot\|_\infty$ formulation as the 
canonical metric for the continuous theory.

\begin{remark}[Practical use of cosine distance]
While our theoretical framework is formulated in $C^0([0,T])$ with the 
supremum norm $\|\cdot\|_\infty$, practical computation requires discretizing 
signals into $\mathbb{R}^N$. In this finite-dimensional setting, we use cosine 
distance on $\ell_2$-normalized vectors for three reasons:

\begin{enumerate}
    \item \textbf{Computational efficiency:} cosine distance is optimally 
    supported by large-scale vector indexes (e.g., FAISS) and approximate 
    nearest-neighbor search.
    \item \textbf{Amplitude invariance:} normalization removes global scaling, 
    which is desirable in sensor-driven domains where amplitude drift is common.
    \item \textbf{Geometric consistency:} normalized signals lie on the unit 
    sphere, where cosine distance corresponds to angular separation and yields 
    stable geometric behaviour.
\end{enumerate}

In finite dimensions, all norms are equivalent 
\cite[Ch.~1]{rudin1991functional}, so compactness, convergence, and the 
existence of a finite perceptual radius are preserved under cosine distance, 
although the numerical value of the radius may differ between metrics.

Empirically, we observe that the key qualitative predictions of the framework—
radius saturation, manifold compactness, and cross-domain consistency—remain 
robust under cosine distance. This validates the use of cosine geometry as a 
faithful practical implementation of the $C^0$ theory.
\end{remark}

\subsection{Practical implementation: incremental radius estimation}

In practice, the perceptual radius can be estimated incrementally as new
realizations are observed. After selecting an initial reference trace $x_0$,
each subsequent signal $x_i$ is embedded as a vector in $\mathbb{R}^N$ and
inserted into a vector index (e.g.\ a FAISS-style nearest-neighbor structure).
The empirical radius is updated online as
\[
\hat r_n = \max_{1 \le i \le n} \|x_i - x_0\|_\infty,
\]
computed either explicitly or through stored distances maintained by the index.

A key practical observation is that the perceptual manifold becomes usable
long before the radius fully converges: the internal structure (cluster
stability, invariants, neighborhood relations) stabilizes early, while late
samples primarily refine the outer boundary. Thus, anomaly detection,
compatibility scoring, and geometric clustering can be deployed immediately,
even when the supremum of the manifold has not yet been fully explored.

This incremental process reflects the self-supervised nature of perceptual
structure: the observer expands its approximation of $\mathcal{M}$ simply by
accumulating realizations, without labels or a predefined model of the
underlying physics.

A practical question now arises: \emph{how does an observer determine when the
perceptual manifold has been fully explored?}  
Although the radius $r$ is mathematically well-defined, in real-world settings 
it must be inferred progressively as new realizations are observed. At early 
stages, observations remain tightly clustered around the reference $x_0$, 
yielding a small empirical radius. As sampling becomes denser, previously unseen 
regions of the manifold appear and the estimated radius expands. Eventually, the 
process saturates: additional samples lie strictly within the existing boundary, 
indicating that the manifold has been completely discovered.

Figure~\ref{fig:manifold_saturation} illustrates this progression. The empirical 
radius $\hat r_n$ grows rapidly at first and then stabilizes once the extremal 
variations of the phenomenon have been observed. This behavior provides a simple 
operational criterion for manifold completion.

To formalize this estimation procedure, we compute $\hat r_n$ incrementally as 
new samples arrive. The pseudocode below summarizes the algorithm used in all 
experiments, implementing the Monte Carlo estimator of the perceptual radius 
and revealing its convergence as sampling becomes dense in the 
state--condition space.
\begin{figure}[h!]
\centering
\begin{tikzpicture}[scale=1.5]
\begin{scope}[shift={(0,0)}]
  \node at (1.2,2.2) {\small Early samples};

  \draw[black] (0,0) rectangle (2,2);

  \draw[blue!40, dashed] (1,1) circle (0.4);

  \foreach \x/\y in {1.05/1.1, 0.95/0.9, 1.0/1.0}{
      \fill[blue!70] (\x,\y) circle (2pt);
  }
\end{scope}

\begin{scope}[shift={(3,0)}]
  \node at (1.2,2.2) {\small Mid-stage (densification)};

  \draw[black] (0,0) rectangle (2,2);

  \draw[blue!40, dashed] (1,1) circle (0.9);

  \foreach \x/\y in {
    0.6/0.9, 1.4/1.1, 1.2/0.6, 0.8/1.5,
    1.6/1.4, 1.0/0.4, 0.7/0.6, 1.3/1.3
  }{
      \fill[blue!70] (\x,\y) circle (2pt);
  }
\end{scope}

\begin{scope}[shift={(6,0)}]
  \node at (1.2,2.2) {\small Saturation of manifold};

  \draw[black] (0,0) rectangle (2,2);

  \draw[blue!70, thick] (1,1) circle (1.1);

  \foreach \x/\y in {
    0.6/0.9, 1.4/1.1, 1.2/0.6, 0.8/1.5,
    1.6/1.4, 1.0/0.4, 0.7/0.6, 1.3/1.3,
    0.9/1.6, 1.5/0.8, 1.1/0.5, 0.5/1.1
  }{
      \fill[blue!70] (\x,\y) circle (2pt);
  }
\end{scope}

\end{tikzpicture}

\caption{\small  
Evolution of the empirical perceptual radius: early samples explore a small  
region, mid-stage sampling expands the estimated radius, and saturation occurs  
when additional observations no longer increase the empirical supremum  
distance. This illustrates the convergence of \(\hat r_n \rightarrow r\).}
\label{fig:manifold_saturation}
\end{figure}

\begin{algorithm}[H]
\caption{Incremental Estimation of the Perceptual Radius}
\begin{algorithmic}[1]

\Require Stream of realizations $(x_1, x_2, \dots)$, reference $x_0$
\State Initialize vector index $\mathcal{I} \gets \emptyset$
\State $\hat r_0 \gets 0$

\For{$n = 1,2,\dots$}

    \State Insert $x_n$ into index: $\mathcal{I} \gets \mathcal{I} \cup \{x_n\}$
    
    \State Compute distance to reference:
    \[
       d_n = \|x_n - x_0\|_\infty
    \]
    
    \State Update empirical radius:
    \[
        \hat r_n = \max(\hat r_{n-1}, d_n)
    \]
    
    \State Optionally return early-warning signals:
    \[
    \text{if } d_n > \hat r_{n-1} + \delta 
       \Rightarrow \text{new variability detected}
    \]
    
\EndFor

\State \Return $\hat r_n$, stabilized perceptual manifold $\mathcal{M}_n$
\end{algorithmic}
\end{algorithm}

While the geometric panels in Figure~\ref{fig:manifold_saturation} convey the
intuition of manifold discovery, the real operational signal of convergence
comes from the evolution of the empirical radius $\hat r_n$ as a function of the
number of observed realizations. 

In deterministic physical systems, new samples initially reveal previously
unseen variability, causing $\hat r_n$ to grow rapidly. However, once the
extremal behaviors of the phenomenon have been observed, the radius enters a
plateau regime: additional realizations remain strictly within the established
boundary, and $\hat r_n$ stabilizes. 

This saturation behavior is the empirical signature of manifold completion.  
It provides a practical, data-driven criterion for determining when the
observer has fully discovered the admissible set of realizations, even without
access to the underlying physical equations. Importantly, saturation does not
mean that sampling stops being useful—internal structure (neighborhoods,
invariants, cluster geometry) stabilizes much earlier—but it marks the point
where the outer boundary of the perceptual manifold has been reached.

The next figure shows a typical saturation curve observed across all datasets:
a sharp initial expansion followed by a gradual flattening toward a stable
limit. This empirical pattern mirrors the theoretical convergence
$\hat r_n \to r$ established in Section~4 and underpins the self-supervised
nature of perceptual discovery in real-world systems.

\begin{figure}[h!]
\centering
\begin{tikzpicture}
\begin{axis}[
    width=10cm,
    height=6cm,
    xlabel={Number of samples $n$},
    ylabel={Empirical radius $\hat r_n$},
    xmin=0, xmax=100,
    ymin=0, ymax=1.2,
    smooth,
    thick,
    grid=major,
    grid style={dashed,gray!30},
    xlabel style={font=\small},
    ylabel style={font=\small},
    ticklabel style={font=\small}
]

\addplot[blue!70, thick]
    table[row sep=\\]{
    x   y
    0   0.00 \\
    5   0.03 \\
    10  0.09 \\
    15  0.22 \\
    20  0.43 \\
    25  0.62 \\
    30  0.73 \\
    40  0.79 \\
    50  0.805 \\
    60  0.812 \\
    70  0.810 \\
    80  0.816 \\
    90  0.814 \\
    100 0.812 \\
};

\end{axis}
\end{tikzpicture}
\caption{\small
Conceptual saturation curve of the empirical perceptual radius.
Early samples reveal local variability, intermediate samples discover the
outer boundary of the manifold, and later samples primarily densify the
already discovered region rather than increasing its empirical radius.
This pattern motivates the operational criterion used in the experiments:
saturation occurs when additional realizations no longer expand the observed
boundary of the perceptual manifold.
}
\label{fig:radius_saturation_curve}
\end{figure}

\begin{remark}[World-evidence and saturation behaviour]
Across all five domains—electromechanical point machines, electrochemical 
battery discharge, physiological ECG morphology, solar irradiance and tides—the empirical evolution of 
the perceptual radius exhibits the same characteristic pattern: a rapid initial 
expansion followed by a clear saturation plateau. This behaviour is not an 
artifact of the metric or the sampling procedure; it reflects a deeper 
physical constraint. Real phenomena do not explore the full geometric extent 
permitted by the space (e.g.\ cosine distance approaching~1). Instead, 
world-evidence constrains variability to a compact region shaped by the 
underlying physics.

Anomalous or non-physical signals can approach the theoretical maximum 
(e.g.\ cosine distance near~1), but nominal realizations never do. Across all 
datasets analysed, the empirical radius stabilizes well below this limit, 
indicating that the admissible perceptual manifold occupies only a bounded, 
low-variability subset of the functional space. This saturation is therefore a 
direct empirical signature of compactness, confirming the theoretical 
predictions of Sections~2--4.
\end{remark}

\subsection{Cross-domain evaluation}

The same pipeline\ref{sec:methods} is applied to the railway point machine dataset, the NASA 
battery aging dataset, the MIT-BIH ECG database, NSRDB solar irradiance data, and NOAA tidal records. 
Because all domains are processed identically, differences in perceptual 
geometry reflect the underlying physical processes rather than methodological 
bias.

Complete implementation details, including preprocessing scripts, Hausdorff
computations, Monte Carlo sampling, and all experimental code, are provided in
the public repository associated with this work:
\url{https://github.com/eduardodisanti/the_geometry_of_intelligence}.

\subsection{Public Datasets Used}

To demonstrate that deterministic functional topology is a general property of 
real-world physical systems, we evaluate our framework across five public 
datasets spanning electromechanical, electrochemical, physiological, atmospheric, and geophysical  
domains.

\subsubsection{Railway Point Machine Current Traces}

We use the public Chinese Railway Point Machine dataset 
\cite{li2020pointmachine, CRPM_dataset_url, KaggleSwitchDataset}.  
The signals exhibit a characteristic deterministic structure (inrush, plateau, 
closure peak), ideal for testing compactness and Hausdorff radius.

\subsubsection{NASA Battery Aging Dataset}

We use the NASA Ames Battery Dataset 
\cite{NASA_Battery_Repository, saha2007prognostics, saha2011modeling}.  
Battery discharge curves are smooth, bounded, and deterministic.

\subsubsection{MIT-BIH Electrocardiogram Dataset}

We use the MIT-BIH Arrhythmia dataset 
\cite{MITBIH_dataset_1980, MITBIH_url, physionet2000}.  
ECG morphology (P, QRS, T waves) forms a low-variability compact manifold.

\subsection{Synthetic generators (overview)}

To assess whether manifold saturation arises from intrinsic physical 
constraints rather than from dataset-specific properties, we complement the 
real signals with deterministic synthetic generators in each domain.  
Each generator produces a compact family of continuous curves constructed from 
simple, domain-agnostic equations with bounded parameters.  
The goal is not to reproduce detailed physics but to create controlled 
functional manifolds whose geometry can be compared directly with that of the 
real systems.

All generators share the same structural design: a fast onset regime, a 
quasi-stationary middle phase, and a terminal transient, with small smooth 
perturbations added to account for structured variability while preserving 
continuity.  
Their explicit equations and parameter ranges are provided in 
Appendices~\ref{app:gen_electromechanical}--\ref{app:gen_ecg}.  
These deterministic manifolds allow us to evaluate the Monte Carlo radius 
estimator under fully controlled conditions and to compare synthetic and 
real-world perceptual geometry on equal footing.

\section{Results: Geometry Across Domains}

\subsection{Point Machines: Functional Manifold Geometry}

We evaluate whether the instantaneous power envelopes of electromechanical
railway point machines (PMs) form a compact functional manifold and whether
Monte Carlo (MC) simulation can approximate its geometry in the absence of
large real datasets. All signals were resampled to 160 points and normalized
under cosine geometry, which induces the same topology as $\|\cdot\|_\infty$
on the finite-dimensional embedding.

Our analysis proceeds in two stages:
\begin{enumerate}
    \item intrinsic saturation of the real PM manifold,
    \item intrinsic saturation of the simulated manifold,
\end{enumerate}

\subsubsection{Saturation of the Real PM Manifold}

For a subset $X_n$ of $n$=8788 real signals, we compute:
\[
d_H(X_n, X_{n/2}), \qquad
r_{\max}(X_n), \qquad
\bar{r}(X_n), \qquad
V_{\mathrm{bbox}}(X_n),
\]
where $d_H(X_n, X_{n/2})$ measures internal geometric stability rather than
distance between distinct physical manifolds.

A striking observation is that \emph{all metrics saturate extremely early}.  
Between $n = 20$ and $50$, the manifold geometry becomes stable:
\[
d_H(X_n, X_{n/2}) \approx 10^{-2}, \qquad
r_{\max}(X_n) \approx \text{constant}, \qquad
V_{\mathrm{bbox}}(X_n) \approx \text{constant}.
\]

This indicates that PM power signals inhabit a
\emph{compact, low-variability functional manifold} shaped almost entirely by
physical constraints (motor torque, inertia, switch mechanism friction, and
closure impact).

\begin{figure}[h!]
\centering
\includegraphics[width=0.90\textwidth]{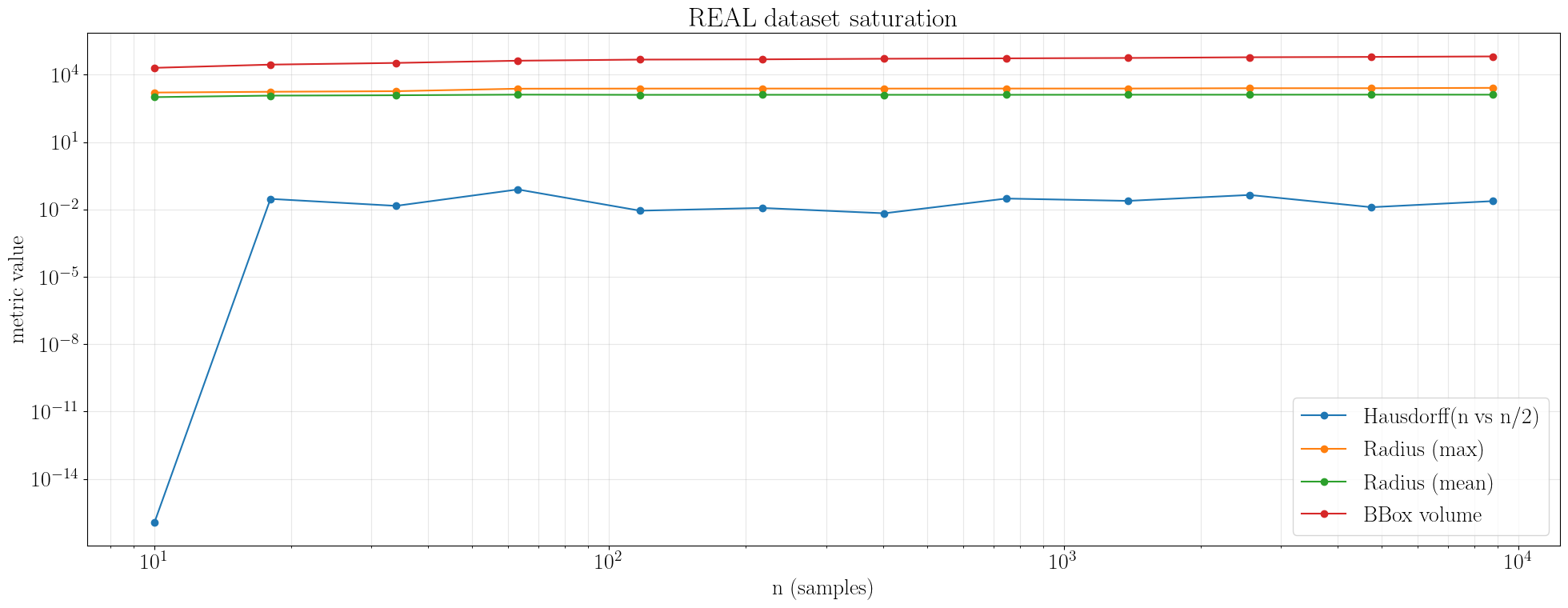}
\caption{\small
Saturation of the real point-machine manifold. All geometric metrics
stabilize after $\sim 20$--$50$ samples, indicating compactness and finite
functional variability.}
\label{fig:pm_real_saturation}
\end{figure}

\subsubsection{Saturation of the Simulated PM Manifold}

We generated 8\,000 Monte Carlo waveforms using a physics-aware AC model.
Despite amplitude and noise differences, the simulated manifold exhibits
\emph{identical saturation behaviour}:
\[
d_H(X^{\mathrm{sim}}_n, X^{\mathrm{sim}}_{n/2}) \approx 10^{-2},
\qquad r_{\max}(X^{\mathrm{sim}}_n) \approx \text{constant}.
\]

The simulator therefore produces signals lying on a compact functional manifold
with the same qualitative saturation signature observed in the real PM dataset,
suggesting that both are governed by similarly bounded functional variability under actuation constraints.

\begin{figure}[h!]
\centering
\includegraphics[width=0.90\textwidth]{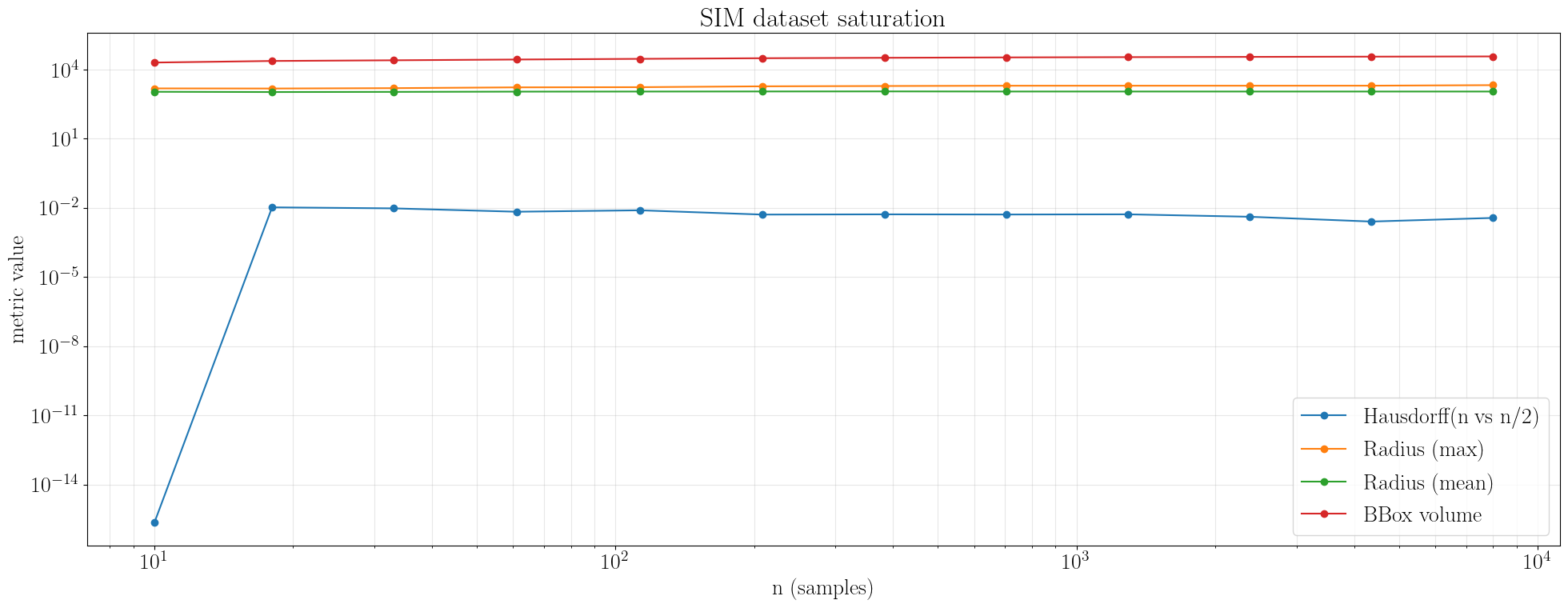}
\caption{\small
Saturation of the simulated point-machine manifold. 
The geometry is stable and compact and exhibits the same qualitative saturation constraints as the real dataset.}
\label{fig:pm_sim_saturation}
\end{figure}

\subsubsection{Waveform Morphology and Simulation Coverage}

\begin{figure}[h!]
\centering
\includegraphics[width=0.90\textwidth]{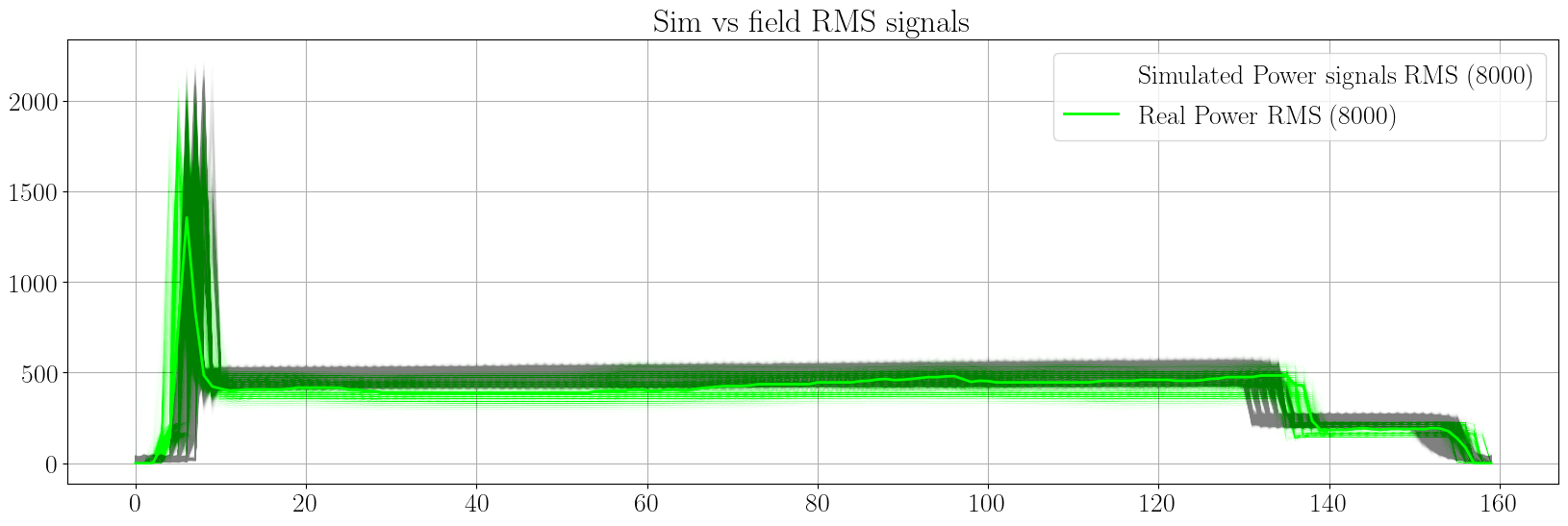}
\caption{\small
Representative RMS instantaneous power envelopes for railway point machines.
In green: randomly sampled real manoeuvres from an in-service machine, showing the
idle--inrush--locking--idle organisation with a terminal impact transient.
In grey: Monte Carlo waveforms generated by the physics-aware AC simulator,
which reproduce the same qualitative morphology while exploring admissible
parametric variability.}
\label{fig:pm_waveforms}
\end{figure}

\subsubsection{Summary of Findings}

Across both real and simulated PM datasets:
\begin{itemize}
    \item Manifolds are compact and low-dimensional.
    \item Saturation occurs with fewer than 50 samples.
    \item Real and simulated manifolds exhibit the same qualitative saturation regime.
    \item MC simulation exhibits the same saturation signature
\end{itemize}

Together, these results confirm that PM power signals form a deterministic,
physically constrained functional manifold whose radius and boundaries can be
estimated reliably using Monte Carlo simulation.

\subsection{Batteries: Electrochemical Discharge Manifolds}

Battery discharge curves from the NASA Ames Prognostics Center of Excellence (PCoE)
lithium-ion battery dataset exhibit smooth, deterministic voltage trajectories
governed by electrochemical kinetics, diffusion processes, and internal resistance
effects. We analyzed a total of $n = 2{,}794$ individual discharge cycles collected
from $34$ commercial lithium-ion cells (B0005--B0056), spanning early-life to
significantly aged operating regimes.

Each discharge corresponds to a constant-current load, with currents ranging
approximately between $2$~A and $4$~A depending on the experimental protocol.
While the nominal capacity of each cell decreases gradually over hundreds of
cycles due to aging, each individual discharge curve represents a physically
constrained realization of the underlying electrochemical system at a given
degradation state.

Despite long-term non-stationarity across cycles, the instantaneous discharge
manifold at any fixed aging stage remains compact. The voltage profile $V(t)$
exhibits a characteristic morphology shared across cells: an initial transient,
a quasi-linear plateau regime, a nonlinear decay phase, and a sharp terminal
cutoff associated with lithium depletion and increased internal resistance.

\subsubsection{Saturation of the Battery Manifold}

Geometric metrics were computed over subsets $X_n$ of discharge curves with
increasing sample size. We observe rapid saturation of all metrics:
\[
d_H(X_n, X_{n/2}) \approx 10^{-2}, \qquad
r_{\max}(X_n) \approx \text{constant}
\quad \text{for } n \gtrsim 50\text{--}100.
\]

\begin{figure}[h!]
\centering
\includegraphics[width=0.90\textwidth]{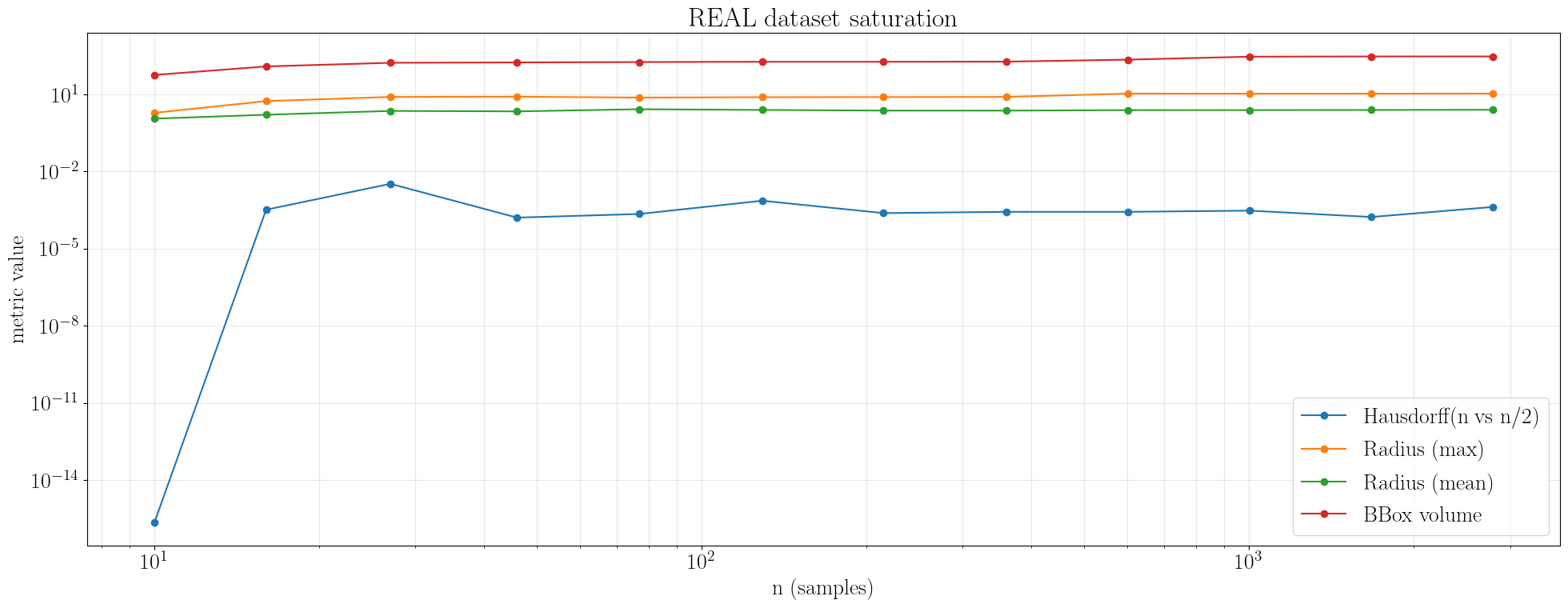}
\caption{\small
Saturation of the battery discharge manifold. The Hausdorff distance between
$X_n$ and $X_{n/2}$, as well as global radius estimates, stabilize rapidly,
indicating that additional discharge curves do not introduce new geometric
structure beyond a small number of samples.
}
\label{fig:batteries_real_saturation}
\end{figure}

Importantly, this saturation is observed despite the \emph{non-stationary} nature
of battery aging across cycles. The result indicates that while aging induces a
slow drift of the discharge manifold over time, the set of physically realizable
voltage trajectories at any given cycle remains strongly constrained and compact.

\subsubsection{Physical Interpretation}
The observed geometric saturation reflects bounded electrochemical variability.
Manufacturing tolerances in electrode composition, small variations in ambient
temperature, measurement noise, and stochastic effects in lithium transport
introduce structured but finite deviations from the nominal discharge trajectory.

From a geometric perspective, the perceptual radius captures the envelope of all
physically admissible discharge profiles under constant-current operation. Aging
acts primarily as a slow deformation of this manifold rather than an expansion
of its intrinsic dimensionality, explaining the early saturation observed across
cycles.


\subsection{ECG: Physiological Heartbeat Morphology}

Electrocardiogram (ECG) signals provide a canonical example of a deterministic
biological process governed by strong physical and physiological constraints.
The morphology of the cardiac cycle, and in particular the QRS complex
associated with ventricular depolarization, is tightly regulated by cardiac
conduction pathways and ionic dynamics. As a result, normal ECG waveforms
exhibit highly reproducible structure with bounded variability.

We analyze normal sinus beats (annotation label \texttt{N}) from the MIT-BIH
Arrhythmia database. Each beat is aligned to the R-peak and resampled onto a
uniform 160-point grid spanning a fixed temporal window around the peak. This
yields a collection of continuous signals embedded in $\mathbb{R}^{160}$ and
treated identically to the other domains studied in this work.

\subsubsection{Saturation of the Real ECG Manifold}

We first examine the intrinsic geometry of the real ECG perceptual manifold.
For increasing subsets $X_n$ of normal beats, we compute internal Hausdorff
stability and extremal radius metrics:
\[
d_H(X_n, X_{n/2}), \qquad r_{\max}(X_n), \qquad \bar{r}(X_n).
\]

The ECG manifold exhibits extremely rapid saturation:
\[
d_H(X_n, X_{n/2}) \approx 10^{-2}, \qquad
r_{\max}(X_n) \approx \text{constant}
\quad \text{for } n \gtrsim 20\text{--}40.
\]

This indicates that the admissible space of normal ECG morphologies is highly
compact and low-dimensional. After a small number of realizations, additional
samples no longer expand the outer boundary of the manifold but merely densify
its interior.

\begin{figure}[h!]
\centering
\includegraphics[width=0.90\textwidth]{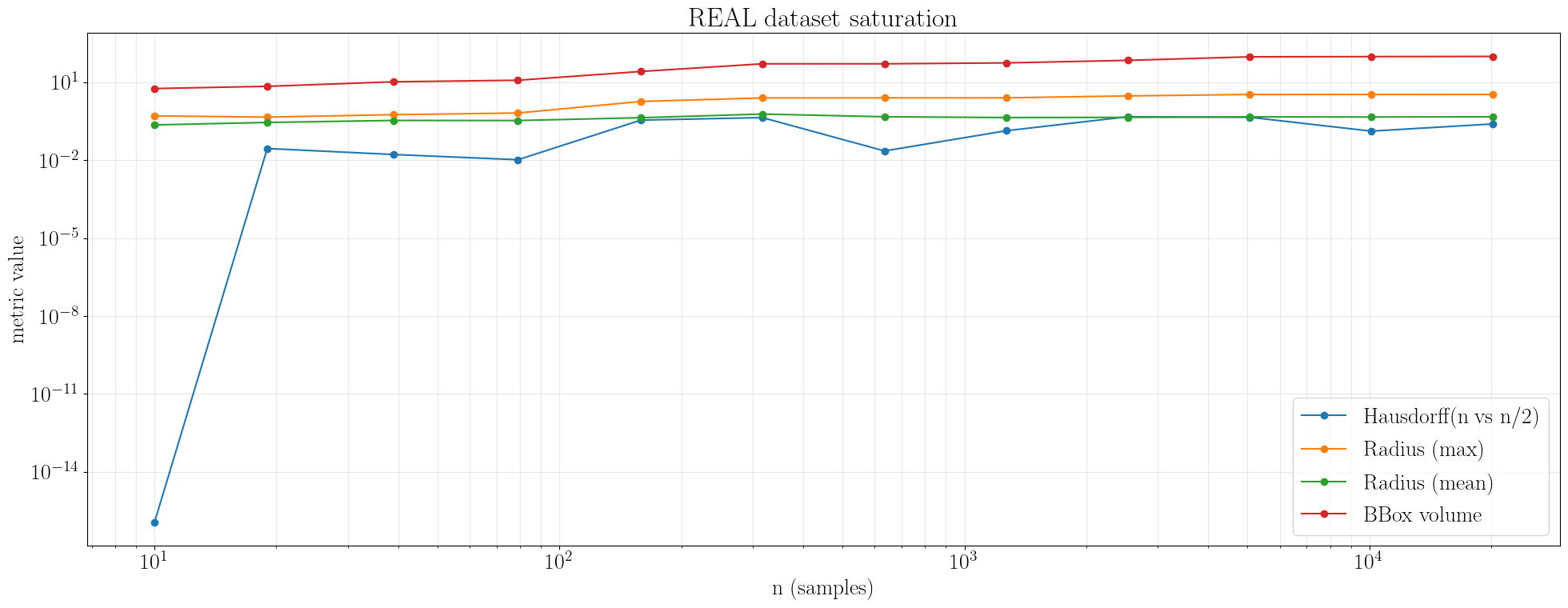}
\caption{\small
Saturation of the real ECG perceptual manifold. Geometric metrics stabilize
after approximately $20$--$40$ samples, demonstrating extreme compactness of
physiological signal spaces.
}
\label{fig:ecg_real_saturation}
\end{figure}

\subsubsection{Synthetic ECG Generators}

To assess whether geometric saturation depends on accurate physiological
modeling, we analyze two synthetic ECG generators with markedly different
levels of realism. These generators are not intended to faithfully reproduce
true cardiac dynamics, but to generate bounded families of continuous waveforms
with controlled variability.

\paragraph{McSharry dynamical generator.}
We consider the well-known low-dimensional ECG model proposed by McSharry
et al.~\cite{mcsharry2003dynamical}, which generates ECG-like signals through a
nonlinear dynamical system designed to approximate the P--QRS--T morphology.
While the resulting waveforms are visually recognizable, they differ
substantially from real ECG signals in fine temporal structure, smoothness, and
relative amplitude balance.

\begin{figure}[h!]
\centering
\includegraphics[width=0.95\textwidth]{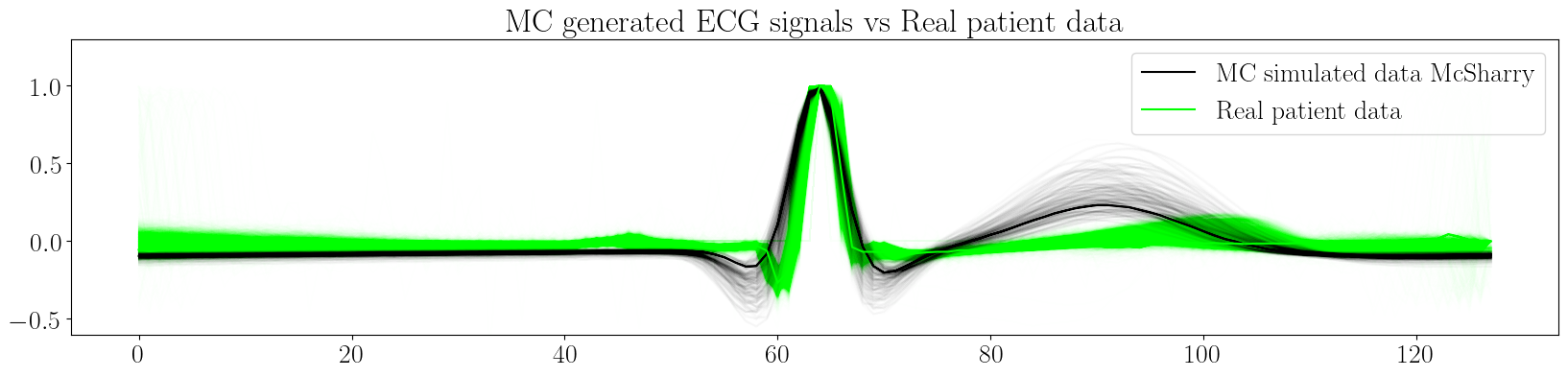}
\caption{\small
Synthetic beats generated by the McSharry dynamical model.
}
\label{fig:ecg_mchsarry_generator_examples}
\end{figure}

\paragraph{Gaussian morphological emulator.}
As a deliberately simplified baseline, we also construct a purely morphological
ECG emulator based on a superposition of Gaussian components representing the
P, QRS, and T waves, with bounded parameters and smooth perturbations. This
generator ignores electrophysiological dynamics entirely and produces visibly
idealized waveforms, yet defines a compact family of continuous signals by
construction.

\begin{figure}[h!]
\centering
\includegraphics[width=0.95\textwidth]{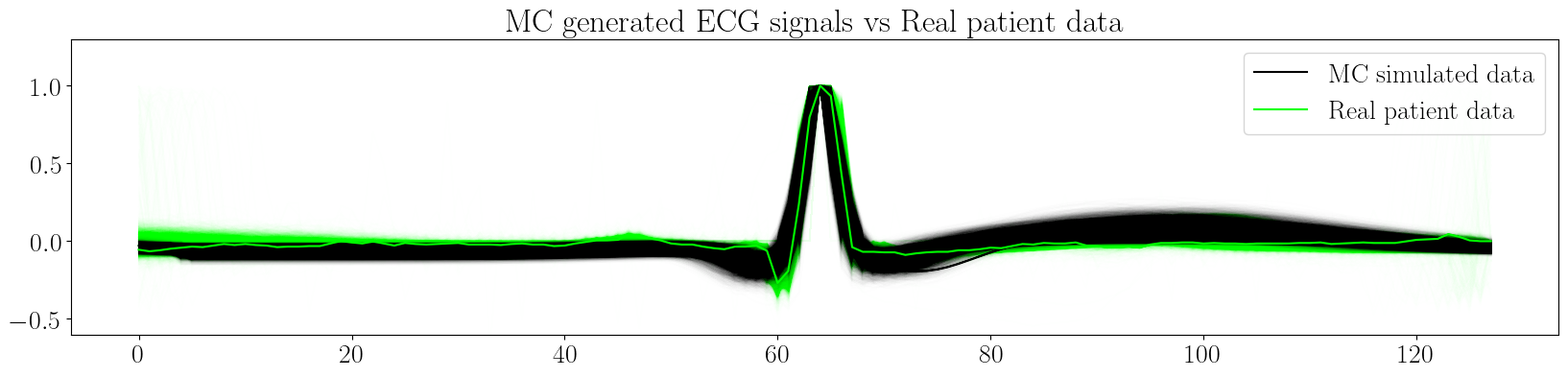}
\caption{\small
Synthetic beats produced by the Gaussian morphological emulator.
}
\label{fig:ecg_gaussian_generator_examples}
\end{figure}
\begin{remark}
The synthetic generators are visibly imperfect and do not reproduce fine physiological detail, yet produce bounded
families of continuous waveforms.
\end{remark}

\subsubsection{Saturation of the McSharry ECG Manifold}

We apply the same Monte Carlo radius estimation pipeline to the ECG signals
generated by the McSharry model. For increasing subsets
$X_n^{\mathrm{McS}}$, we compute internal stability and extremal radius metrics.

Despite the clear morphological discrepancies with real ECG recordings, the
synthetic manifold exhibits rapid geometric saturation:
\[
d_H(X_n^{\mathrm{McS}}, X_{n/2}^{\mathrm{McS}}) \approx 10^{-2}, \qquad
r_{\max}(X_n^{\mathrm{McS}}) \approx \text{constant}
\quad \text{for } n \gtrsim 30\text{--}50.
\]

This shows that accurate physiological realism is not required for saturation
to emerge. The McSharry generator produces a compact functional manifold with
bounded variability, and Monte Carlo sampling quickly exhausts its admissible
space.

\begin{figure}[h!]
\centering
\includegraphics[width=0.90\textwidth]{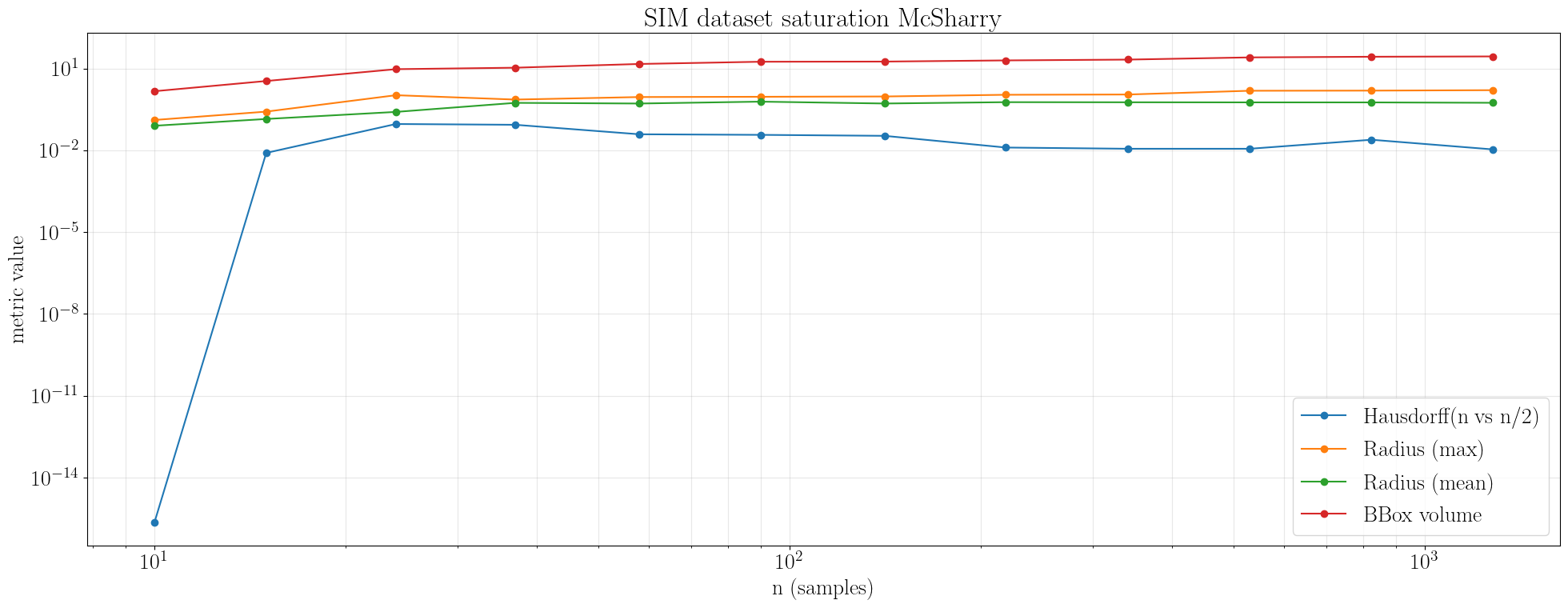}
\caption{\small
Saturation of the ECG manifold generated by the McSharry dynamical model.
Geometric metrics stabilize after a small number of samples, indicating a
compact functional manifold despite imperfect physiological realism.
}
\label{fig:ecg_mcsharry_saturation}
\end{figure}

\subsubsection{Saturation of the Gaussian ECG Manifold}

We repeat the same analysis for the Gaussian morphological emulator. Although
this generator lacks any electrophysiological or dynamical grounding, its
parameters are bounded and the resulting signals are continuous by design.

The empirical perceptual radius again saturates extremely rapidly:
\[
d_H(X_n^{\mathrm{Gauss}}, X_{n/2}^{\mathrm{Gauss}}) \approx 10^{-2}, \qquad
r_{\max}(X_n^{\mathrm{Gauss}}) \approx \text{constant}
\quad \text{for } n \gtrsim 20\text{--}40.
\]

This confirms that saturation is a geometric consequence of bounded
deterministic variability rather than a byproduct of detailed physical
modeling or simulator fidelity.

\begin{figure}[h!]
\centering
\includegraphics[width=0.90\textwidth]{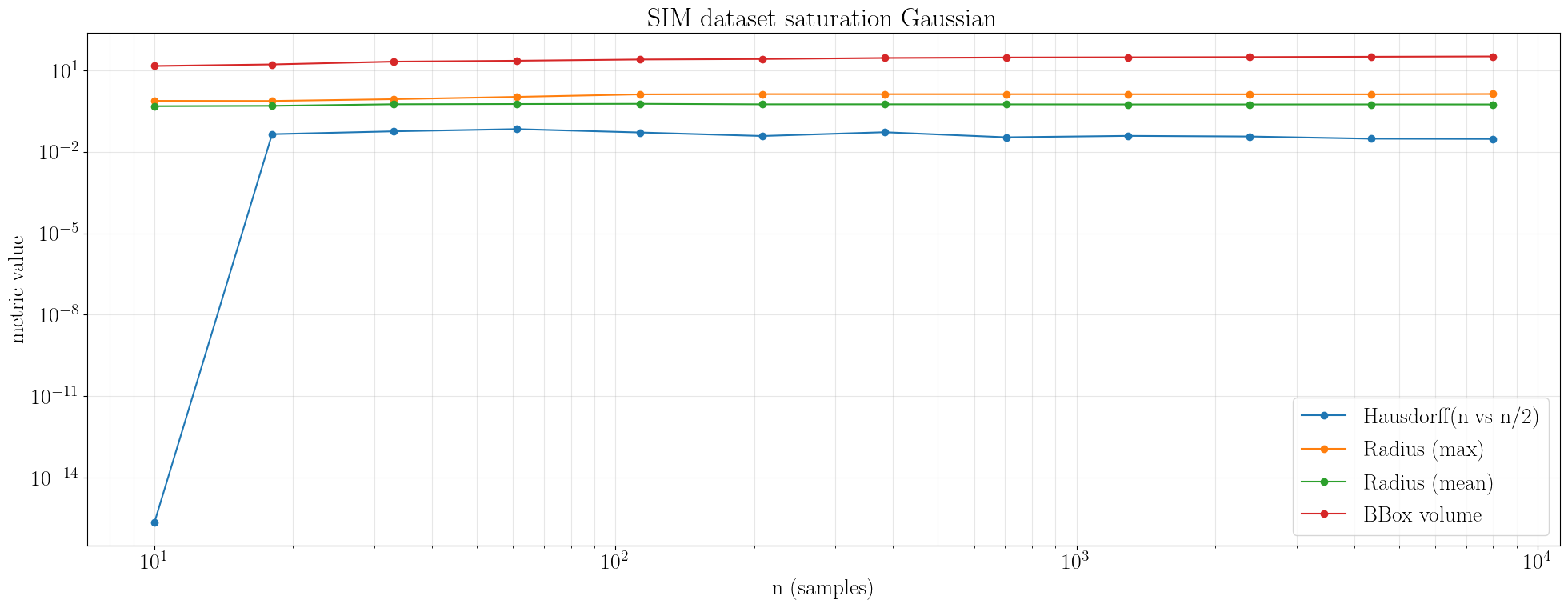}
\caption{\small
Saturation of the ECG manifold generated by a Gaussian morphological emulator.
Even with a highly simplified and non-physiological generator, the perceptual
manifold remains compact and exhibits rapid radius convergence.
}
\label{fig:ecg_gaussian_saturation}
\end{figure}

\subsubsection{Summary: Saturation Across Real and Synthetic ECG Manifolds}

Across real ECG recordings, the McSharry dynamical generator, and the simplified
Gaussian morphological emulator, we observe the same qualitative geometric
behavior:
\begin{itemize}
    \item All ECG manifolds are compact and low-dimensional.
    \item The perceptual radius saturates after a small number of samples.
    \item Internal geometric stability emerges early and persists with
    increasing sampling.
\end{itemize}

Crucially, saturation occurs even when the synthetic generators fail to
accurately reproduce fine physiological details. This demonstrates that
geometric compactness and rapid saturation are properties of bounded
deterministic signal families, not of simulator fidelity. The role of learning
or simulation is therefore to explore the admissible perceptual manifold rather
than to perfectly replicate the underlying physical process.

\subsection{NSRDB multi-site solar manifolds}

To test whether rapid geometric saturation also appears in an
atmospheric domain with strong seasonal structure, we considered
hourly global horizontal irradiance (GHI) data from the NOAA/NCEI
National Solar Radiation Database (NSRDB). We sampled 100 stations at
random from the public \emph{solar-only} archive and, for each site,
constructed a daily irradiance manifold using the same pipeline as in
the previous domains. After removing invalid entries and resolving
timestamps of the form \texttt{24:00} as \texttt{00:00} on the following
day, we grouped the GHI time series by calendar day, reindexed each day
to a fixed 24-hour local-time grid, and normalized each daily curve by
its daily maximum. Each day was therefore represented as a vector in
$\mathbb{R}^{24}$, and the collection of such vectors defined the
station-specific solar manifold. 

We then applied exactly the same saturation analysis used throughout the
paper: as the number of sampled daily profiles increased, we computed
the internal Hausdorff discrepancy, the maximum radius, the mean radius,
and the bounding-box volume. To summarize the onset of saturation across
stations, we defined an operational saturation scale
$n_{\mathrm{sat}}$ from the mean-radius curve as the first sample size
for which the relative change between successive steps fell below 5\%
for two consecutive steps. Across the random station sample, the median
estimated saturation scale was 22 days, with 68.75\% of stations
saturating by 30 days and 81.25\% by 40 days. This indicates that, at a
fixed site, the manifold of normalized daily irradiance profiles is
typically resolved after only a few tens of days.

Representative saturation curves for randomly selected stations are
shown in Appendix~\ref{app:nsrdb_multisite}.

\subsection{NOAA multi-site tidal manifolds}

To examine the limits of geometric saturation in a purely geophysical 
domain governed by orbital mechanics, we analyzed sea-level variations 
from the NOAA CO-OPS database\cite{noaa_coops_tidescurrents}. We selected 9 stations at random 
representing diverse coastal environments—including the Pacific, 
Atlantic, and Gulf coasts—and constructed station-specific tidal 
manifolds. Following our functional pipeline, we segmented the water-level 
time series into daily 24-hour cycles, reindexed them to a fixed hourly 
grid, and applied a min-max normalization. This process isolates the 
characteristic "signature" of the local tidal harmonics from absolute 
mean sea-level shifts, representing each day as a stable topological 
realization in $\mathbb{R}^{24}$.

The saturation analysis reveals that the tidal manifold is among the 
most compact structures studied. Using the same operational saturation 
scale $n_{\mathrm{sat}}$ (relative change below 5\% for two consecutive 
steps), we observed a median saturation point of only 10 days. 
Remarkably, the entire sample achieved full geometric saturation within 
31 days, a scale that aligns with the fundamental periodicity of the 
lunar synodic month. This indicates that once a system has "experienced" 
 a full cycle of spring and neap tides, the underlying manifold is 
completely resolved. 

The abruptness of this collapse—faster even than in the solar domain—highlights 
how the absence of stochastic atmospheric interference (such as cloud 
cover) allows the deterministic "Blueprint" of the physical process to 
emerge almost instantaneously. Detailed saturation curves and aggregate 
statistics for all sampled tidal stations are provided in 
Appendix~\ref{app:noaa_multisite}.

\subsection{Summary of Results for Deterministic Datasets}

Table~\ref{tab:saturation_summary} provides a comparative overview of the 
operational saturation scale across all physical and physiological domains 
studied. The remarkable consistency of $n_{\text{sat}}$---ranging from 10 to 40 
samples despite the vast differences in underlying physics and sampling 
frequencies---constitutes strong empirical evidence for the existence of 
compact functional "Blueprints" in real-world processes.

\begin{table}[h!]
\centering
\begin{tabular}{@{}lccc@{}}
\toprule
Domain / dataset & Type & $n_{\text{sat}}$ (approx.) & Notes \\
\midrule
Railway point machines (CRPM) & Electromechanical & 15--20 & Real current traces \\
NASA battery aging (PCoE)     & Electrochemical   & 20--30 & Constant-current discharge \\
MIT-BIH ECG (normal beats)    & Physiological     & 30--40 & R-peak aligned beats \\
Synthetic PM generator        & Electromechanical & 20--30 & Bounded parameter family \\
Synthetic ECG generators      & Physiological     & 20--30 & McSharry / Gaussian models \\
NSRDB (100 random sites)      & Solar radiation   & 20--25 & Multi-site saturation \\
NOAA Tides (9 random sites)   & Geophysical       & 10--20 & Gravitational harmonics \\
\bottomrule
\end{tabular}
\caption{\small
  Approximate sample size $n_{\text{sat}}$ at which the empirical
Hausdorff radius stabilizes. In all cases, geometric saturation occurs 
after a surprisingly small number of samples, reflecting the intrinsic 
compactness of the underlying physical manifolds.}
\label{tab:saturation_summary}
\end{table}

\subsection{Preliminary Observations on Perceptual Benchmarks}
\label{sec:results-mnist-spoken}

While our main focus is on physical and physiological processes, it is
natural to ask whether the same geometric signatures---compact manifolds
and early saturation of the empirical radius---also appear in standard
perceptual benchmarks.

To probe this question, we conducted a preliminary geometric analysis on
two canonical datasets:
\begin{itemize}
  \item the MNIST handwritten digit dataset~\cite{lecun1998mnist}, and
  \item a spoken-digit dataset SpeechMNIST~\cite{khacef2020writtenspoken}.
\end{itemize}
In both cases, each class $c \in \{0,\dots,9\}$ is treated as a perceptual
manifold $\mathcal{M}_c$ obtained by embedding the corresponding signals
(images or audio-derived feature trajectories) into $C^0([0,T])$ using the
same preprocessing pipeline as in the physical domains
(Section~\ref{sec:methods}).

\begin{figure}[ht] 
  \centering
  \small
  %
  %
  \begin{subfigure}[b]{0.48\linewidth}
    \centering
    \includegraphics[width=\linewidth]{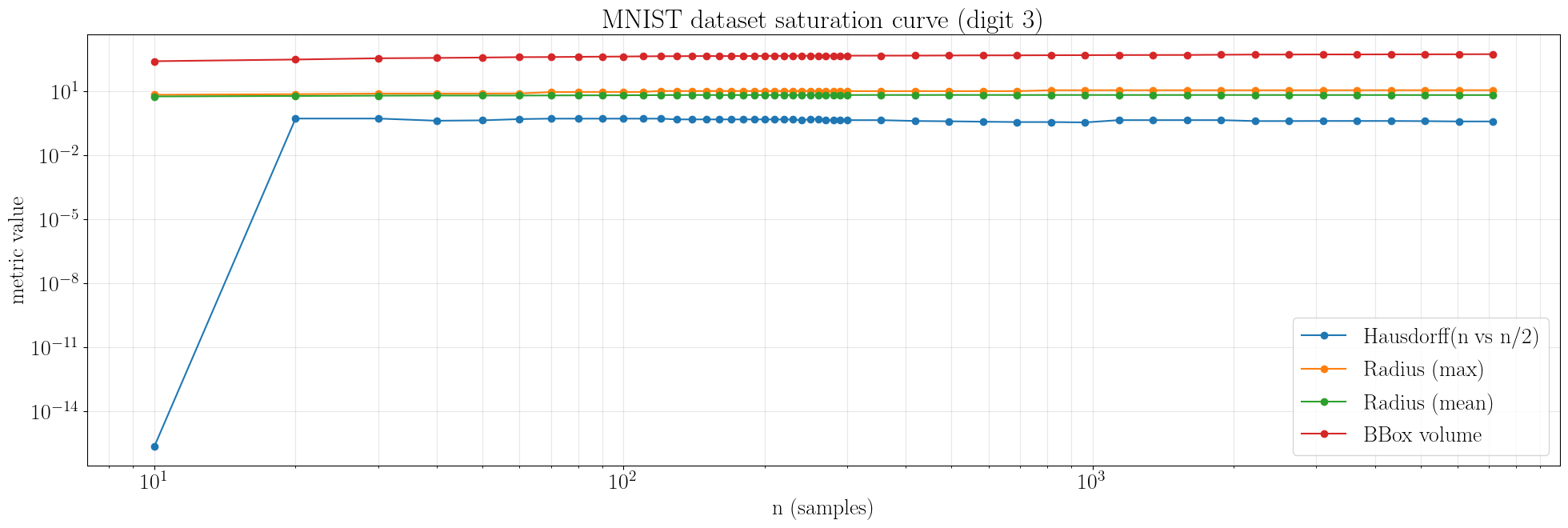}
    \caption{\small MNIST digit ``3''}
    \label{fig:mnist_digit3_radius}
  \end{subfigure}
  \hfill
  %
  %
  \begin{subfigure}[b]{0.48\linewidth}
    \centering
    \includegraphics[width=\linewidth]{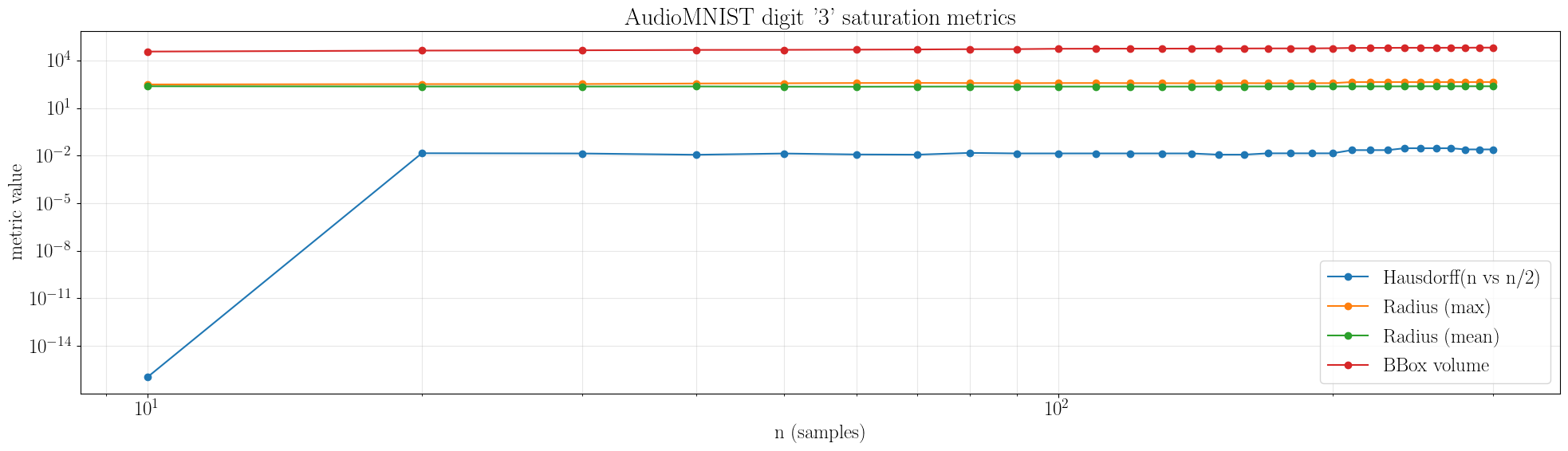}
    \caption{\small Spoken digit ``3''}
    \label{fig:spoken_digit3_radius}
  \end{subfigure}

  \caption{\small
  Empirical Hausdorff radius saturation for the digit ``3'' in MNIST
  (left) and in a spoken-digit dataset (right). In both modalities the
  empirical perceptual radius $\hat r_n$ grows rapidly with the first
  samples and then stabilizes after only a few dozen examples, mirroring
  the behaviour observed in physical and physiological manifolds.}
  \label{fig:mnist_radius_example}
\end{figure}

Figure~\ref{fig:mnist_radius_example} illustrates a typical saturation
curve of the empirical perceptual radius $\hat r_n$ for a single digit
class (here, the digit ``3''): the estimated radius grows quickly as
early samples explore new regions of the manifold, and then stabilizes
after only a few dozen examples. We observe analogous behaviour across
all digit classes and in the spoken-digit dataset, suggesting that the
admissible variability of each concept is bounded and compact in the same
sense as in the physical domains.


Figure~\ref{fig:mnist_digit3_radius} illustrates the empirical
Hausdorff radius saturation for the digit ``3'' in MNIST and in the
spoken-digit dataset. Table~\ref{tab:digit_saturation} reports the corresponding 
saturation sample sizes $n_{\text{sat}}$ for all ten digits in each modality.

\begin{table}[t]
\centering
\begin{tabular}{@{}lcc@{}}
\toprule
Digit & $n_{\text{sat}}$ (MNIST) & $n_{\text{sat}}$ (spoken) \\
\midrule
0 & $\approx 20$ & $\approx 20$ \\
1 & $\approx 20$ & $\approx 20$ \\
2 & $\approx 20$ & $\approx 20$ \\
3 & $\approx 20$ & $\approx 20$ \\
4 & $\approx 20$ & $\approx 20$ \\
5 & $\approx 20$ & $\approx 20$ \\
6 & $\approx 20$ & $\approx 20$ \\
7 & $\approx 20$ & $\approx 20$ \\
8 & $\approx 20$ & $\approx 20$ \\
9 & $\approx 20$ & $\approx 20$ \\
\bottomrule
\end{tabular}
\caption{\small
Approximate saturation sample size $n_{\text{sat}}$ per digit class in
MNIST and in the spoken-digit dataset, using the operational definition
in the main text. In both modalities, the empirical perceptual radius
and internal Hausdorff distance stabilize after roughly twenty examples
per class.}
\label{tab:digit_saturation}
\end{table}

We emphasize that this subsection is purely geometric: we do not specify
a particular classifier here, nor do we aim to exhaustively benchmark
recognition performance. Instead, our goal is to document that the same
early saturation pattern arises in canonical perceptual datasets when
classes are treated as compact manifolds under the proposed framework.
A more systematic study---including comparisons with classical
classifiers (e.g., SVMs) under few-sample regimes—is left for future
work.

\paragraph{While the framework is demonstrated here on physical and physiological signals, its extension to discrete perceptual categories (e.g., computer vision) follows the same principle of manifold compactness, which will be explored in future work focusing on topological classifiers}

\section{Discussion}
\label{sec:discussion}

\subsection{What this work establishes}

The present work makes a concrete claim about deterministic signal
generation: under standard regularity conditions, the set of realizations
produced by a physical process forms a compact perceptual manifold with
finite empirical radius and rapid Hausdorff saturation, on which continuous perceptual functionals are
universally approximable. Our theoretical results formalize this statement
in the Banach space $C^0([0,T])$, provide a Monte Carlo estimator for the
perceptual radius, and give conditions under which the empirical radius
converges as sampling becomes dense in the state--condition space.

Empirically, we show that this compact-manifold picture accurately
describes several real-world signal families. In electromechanical railway
point machines, electrochemical battery discharge trajectories, 
physiological ECG waveforms, solar irradiance profiles and tidal cycles---together with their deterministic
generators---the empirical Hausdorff radius and internal geometric metrics
saturate after surprisingly few samples. Across all these domains, the
admissible signals occupy compact, low-variability regions of functional
space, and the internal geometry stabilizes rapidly once the extremal
realizations have been observed.

\subsection{Self-supervised emergence of perceptual structure}

A central implication of the framework is that perceptual structure can
emerge in a self-supervised manner when explicit physical equations are
unavailable. The perceptual manifold $\mathcal{M}$ and its radius $r$ need
not be specified a priori: they are discovered progressively from the
stream of observations by tracking the evolution of the empirical radius
$\hat r_n$ and related geometric quantities. When $\hat r_n$ (and the
internal Hausdorff distance $d_H(X_n,X_{n/2})$) enters a stable plateau,
the observer has effectively explored the admissible variability of the
process.

In this sense, deterministic physical systems inherently support
self-supervised learning: by repeatedly observing the same process under
varying conditions, an agent can infer the boundaries of the corresponding
perceptual class without labels, external supervision, or a parametric
model. Our experiments in the five physical domains demonstrate this
behaviour in practice, and the preliminary geometric analysis of MNIST and
spoken-digit manifolds in Section~\ref{sec:results-mnist-spoken} suggests
that an analogous mechanism may operate in standard perceptual datasets as
well.

\subsection{Algorithmic and industrial implications}

The compact-manifold viewpoint offers practical advantages for sensing,
diagnosis, and similarity search. In domains such as railway maintenance,
battery monitoring, and physiological signal analysis, the perceptual
manifold provides a compact reference model against which new observations
can be compared. Deviations can then be quantified as distances to the
manifold, enabling interpretable anomaly detection and principled
compatibility scoring. Because the empirical radius typically saturates
after a modest number of samples, such models can be constructed from
relatively small datasets and refined incrementally over time.

Early saturation of the perceptual radius also has direct consequences for
algorithm design. Once the admissible manifold of a deterministic process
is effectively explored, additional subsequences or realizations rarely
expand its radius, suggesting that exhaustive comparison can be replaced by
targeted refinement and early stopping. Recent work (currently under
review) instantiates this idea in the context of matrix-profile--based
analysis of deterministic time series, exploiting radius saturation to
prune redundant comparisons and achieve substantial speedups while
preserving the relevant similarity structure. We view this as a first
example of a broader class of geometric algorithms that operate directly
on compact perceptual manifolds rather than on unconstrained sequence
spaces.

\subsection{Connections to world models and structured representations}

The geometric interpretation aligns with emerging views in machine learning
that emphasize structured representations and world models. A deterministic
physical process implicitly defines a generative mechanism with stable
invariants, finite variability, and predictable boundaries; the associated
perceptual manifold serves as a low-dimensional world model capturing all
admissible observations consistent with the underlying physics.

Unlike traditional latent-variable approaches, however, our framework does
not rely on probabilistic assumptions or explicit parameterizations. The
perceptual manifold is defined by the system itself, and the observer's
task is to approximate its structure with increasing fidelity using
geometric measurements such as the empirical radius and internal
Hausdorff distances. This provides a deterministic counterpart to
world-model learning and suggests a principled way to combine physical
constraints with learned representations in settings where governing
equations are partially known.

Our functional--topological framework is complementary to JEPA-based
architectures such as VL-JEPA~\cite{chen2025vljepa}. While VL-JEPA
empirically shows that predicting target embeddings in a continuous latent
space leads to efficient vision--language learning, our results help
explain why such an approach is plausible in real-world domains:
deterministic physical processes generate compact perceptual manifolds
with finite empirical radius and Hausdorff saturation, on which continuous perceptual functionals
are universally approximable. An interesting direction for future work is
to use manifold-level quantities such as empirical radius and saturation
behaviour to analyse or regularize the embedding spaces learned by
JEPA-style models.

\subsection{Universality as an open hypothesis}

The consistency of our findings across five physical domains, their
deterministic generators, and preliminary analyses of standard perceptual
benchmarks suggests that compact perceptual manifolds may be a widespread
feature of real-world signal families. However, we stress that this should
be regarded as an open hypothesis rather than a universal law. Many
processes of practical interest involve stochastic dynamics, structural
breaks, or evolving constraints that can violate the assumptions of our
framework, and our current experiments do not attempt to cover such cases
systematically.

A systematic characterization of where the compact-manifold picture holds,
how it fails, and how it interacts with modern representation-learning
architectures is an important direction for future work. In this sense,
the present study should be read as a step toward a geometric theory of
perception and representation, not as its completion.

This work represents an initial step in a broader research program aimed at
characterizing the geometric structure of signal manifolds across domains.
Ongoing work is extending the empirical analysis to a wider range of datasets,
including additional physical systems as well as standard perceptual benchmarks
such as MNIST, Fashion-MNIST, CIFAR, and time-series datasets from the UCR/UEA archive.
Preliminary experiments also suggest that similar saturation behavior may
emerge in structured discrete representations, such as molecular fingerprints,
although this setting lies outside the strictly functional framework considered here.

The objective of this program is not to rely on isolated examples, but to
systematically evaluate whether compactness and early geometric saturation
constitute a general structural property of signal families arising from both
physical processes and learned representations.

\section{Conclusion}
\label{sec:conclusion}

This work has developed a deterministic functional–topological framework
for perception and representation, in which the set of realizations
produced by a physical process is modeled as a compact perceptual
manifold in $C^0([0,T])$ with finite Hausdorff radius and stable
invariants. Within this framework, a perceptual category is not an
arbitrary collection of samples, but a compact subset of function space
whose boundaries are dictated by the underlying dynamics and whose
internal variability is inherently limited.

On such manifolds, identification reduces to distance minimization with
respect to the admissible set, and learning corresponds to approximating
a continuous perceptual functional defined on that set. By restricting
attention to compact domains, classical universal approximation results
guarantee that finite-dimensional models can approximate perceptual
functionals arbitrarily well, providing a geometric explanation for the
empirical success of neural networks and related nonlinear architectures
without requiring global probabilistic assumptions.

We have shown that this compact-manifold picture is consistent with the
geometry of several real-world signal families. In electromechanical
railway point machines, electrochemical battery discharge trajectories,
and physiological ECG waveforms—together with their deterministic
generators—the empirical Hausdorff radius and internal geometric metrics
saturate after relatively few samples, indicating that admissible signals
occupy compact, low-variability regions of functional space. Preliminary
analyses of standard perceptual benchmarks (MNIST and spoken digits)
exhibit an analogous early saturation pattern when classes are embedded
as manifolds, suggesting that the same geometric ideas may extend beyond
purely physical processes.

An important consequence of compactness is that perceptual structure can
emerge in a self-supervised manner. When governing equations are unknown
or only partially specified, the perceptual manifold and its radius can
be inferred directly from the stream of observations: as realizations are
accumulated, the empirical radius stabilizes and the admissible set
ceases to expand in any significant way. This offers a principled
geometric foundation for self-supervised learning in real-world systems
and connects deterministic physics with modern representation-learning
paradigms.

Finally, the compact-manifold viewpoint has concrete algorithmic and
practical implications. By exploiting early saturation of the perceptual
radius, one can design observers and similarity-search procedures that
operate on compact reference manifolds rather than on unconstrained
sequence spaces, with applications to sensing, diagnostics, predictive
maintenance, and matrix-profile–based analysis of deterministic time
series. We see this work not as a final theory of the “geometry of
intelligence”, but as a step toward a systematic account of how the
structure imposed by the world shapes the representations that intelligent
systems can and must learn.

\section{Limitations and Future Work}
\label{sec:limitations}

The framework developed in this work is intentionally focused on
deterministic, continuous physical processes, and its scope is
correspondingly limited. Several assumptions underpin our results and
point to natural directions for future research.

First, we assume that the underlying dynamics are effectively
deterministic with bounded noise, so that the set of realizations forms a
compact subset of $C^0([0,T])$. Strongly stochastic systems,
non-stationary regimes, or processes with abrupt structural changes may
violate these assumptions, and our guarantees need not hold in those
settings. Extending the geometric framework to partially deterministic,
regime-switching, or explicitly stochastic systems is an important
direction for future work.

Second, our analysis is restricted to one-dimensional temporal signals
with a fixed observation window. We do not treat higher-dimensional
spatial fields, image sequences, or event-based data in a rigorous way,
although the same functional–topological principles may apply. A careful
study of spatio–temporal manifolds, their associated invariants and
radii, and the role of discretization in such settings remains open.

Third, we work in the topology of $C^0([0,T])$ with the supremum norm and
the induced Hausdorff distance. While this choice is natural for many
sensing applications, other function spaces or metrics may be more
appropriate in different domains (e.g., Sobolev spaces, weighted norms,
or task-dependent distances). A systematic comparison of alternative
topologies and their impact on perceptual geometry is beyond the scope of
this work.

Fourth, our estimation of the perceptual radius from data is necessarily
based on finite sampling. The convergence guarantees rely on increasingly
dense coverage of the state–condition space; in practice, rare operating
regimes, degraded modes, or extreme conditions may be underrepresented.
As a result, empirical estimates of the radius may underestimate the true
variability of the system. Developing adaptive sampling strategies and
explicit coverage criteria would strengthen the practical robustness of
the approach.

Fifth, while we show that perceptual functionals are universally
approximable on compact perceptual manifolds, we do not prescribe a
specific learning algorithm nor provide complexity or sample-efficiency
bounds. Our results are existential rather than algorithmic: they state
that suitable approximators exist, not that any given architecture or
training procedure will find them. Bridging this gap between geometric
existence results and concrete learning algorithms is an important avenue
for future work.

Finally, the present study focuses on the perceptual layer of
intelligence: the acquisition of compact manifolds of admissible
realizations and the definition of a perceptual radius. We do not address
higher-level cognition, hierarchical planning, or decision-making.
Extending deterministic functional topology to multi-layer world models
and control architectures, and combining it with energy-based or
optimization-based inference mechanisms, represents a promising
direction for connecting this framework to fully autonomous agents.
\bibliographystyle{unsrt}
\bibliography{references}

\appendix
\renewcommand{\thesection}{Appendix \Alph{section}}
\section*{Appendices}

\section{Mathematical Proofs}

This appendix provides complete proofs for the theorems and propositions stated
in Sections~2–4. All results are stated in the Banach space 
$C^0([0,T])$ equipped with the supremum norm $\|x\|_\infty$.

\subsection{Proof of Theorem~\ref{thm:compactness} (Compactness of Deterministic Signals)}

\begin{proof}
Let $\mathcal{F} = \{ f(s,\theta) : s\in\mathcal{S},\;\theta\in\Theta\}$ be the
family generating the perceptual set $\mathcal{M}$.  
By assumption:

1. The family is \emph{uniformly bounded}, meaning there exists $M>0$ such that  
   $\|f(s,\theta)\|_\infty \le M$ for all $(s,\theta)$.

2. The family is \emph{equicontinuous}, i.e.\ for every $\varepsilon>0$ there
   exists $\delta>0$ such that for all $t_1,t_2\in[0,T]$:
   \[
   |t_1 - t_2| < \delta
   \quad\Rightarrow\quad
   |f(s,\theta)(t_1) - f(s,\theta)(t_2)| < \varepsilon.
   \]

By the Arzelà--Ascoli Theorem 
\cite{rudin1991functional, royden2010real}, any uniformly bounded and
equicontinuous family of functions has compact closure in $C^0([0,T])$.
Thus $\mathcal{M}$ is compact.
\end{proof}

\subsection{Proof of Proposition~\ref{prop:radius} (Finiteness of the Perceptual Radius)}

\begin{proof}
Because $\mathcal{M}$ is compact in the Banach space $C^0([0,T])$, it is
bounded. Therefore there exists $R > 0$ such that:
\[
\|x\|_\infty \le R
\quad\text{for all } x\in\mathcal{M}.
\]
Fix $x_0\in \mathcal{M}$.  
Then for any $x\in\mathcal{M}$,
\[
\|x - x_0\|_\infty \le \|x\|_\infty + \|x_0\|_\infty \le 2R.
\]
Hence:
\[
r = \sup_{x\in\mathcal{M}} \|x - x_0\|_\infty < \infty.
\]
\end{proof}

\subsection{Proof of Proposition~\ref{prop:uniform} (Uniform Continuity of $\Phi$)}

\begin{proof}
Since $\Phi$ is continuous on $\mathcal{M}$ and $\mathcal{M}$ is compact, 
the classical Heine–Cantor theorem implies that $\Phi$ is uniformly
continuous on $\mathcal{M}$.  
Thus for every $\varepsilon>0$ there exists $\delta>0$ such that
\[
\|x - y\|_\infty < \delta 
\quad\Rightarrow\quad
|\Phi(x) - \Phi(y)| < \varepsilon
\]
for all $x,y\in\mathcal{M}$.
\end{proof}

\subsection{Proof of Theorem~\ref{thm:uat} (Universal Approximation on the Perceptual Manifold)}

\begin{proof}
Let $\mathcal{M} \subset C^0([0,T])$ be compact and let 
$\Phi : \mathcal{M} \to \mathbb{R}$ be continuous.

By Proposition~\ref{prop:uniform}, continuity of $\Phi$ on the compact set 
$\mathcal{M}$ implies uniform continuity. Hence, for every $\varepsilon > 0$ 
there exists $\delta > 0$ such that
\[
\|x - y\|_\infty < \delta
\quad \Rightarrow \quad
|\Phi(x) - \Phi(y)| < \varepsilon
\qquad \text{for all } x,y \in \mathcal{M}.
\]

Because $\mathcal{M}$ is compact in $C^0([0,T])$, it admits a finite 
$\delta$-net with respect to the supremum norm. Equivalently, there exists a 
finite sampling resolution $N$ and a discretization map
\[
\pi_N : \mathcal{M} \to \mathbb{R}^N
\]
such that $\|x - y\|_\infty < \delta$ implies
$\|\pi_N(x) - \pi_N(y)\|_2 < \delta'$ for some $\delta' > 0$, and such that the
sampling error induced by $\pi_N$ remains uniformly below $\delta$ on 
$\mathcal{M}$.

Define the induced mapping
\[
\Phi_N = \Phi \circ \pi_N^{-1}
\]
on the compact set $\pi_N(\mathcal{M}) \subset \mathbb{R}^N$.
By construction, $\Phi_N$ is continuous on a compact subset of a finite-dimensional
Euclidean space.
\paragraph{Remark on scope of approximation.}
The approximation result above does not claim universal approximation over the
entire infinite-dimensional space $C^0([0,T])$. Instead, it applies to continuous
perceptual mappings restricted to compact subsets $\mathcal{M}$ arising from
deterministic physical processes. Compactness implies the existence of finite
$\delta$-nets and, equivalently, finite-dimensional embeddings induced by uniform
sampling. The use of the Universal Approximation Theorem is therefore confined
to these induced finite-dimensional representations, which fully characterize
the perceptual manifold at the chosen resolution and are consistent with all
practical signal processing implementations.

By the Universal Approximation Theorem
\cite{cybenko1989approximation, hornik1991approximation}, for every 
$\varepsilon > 0$ there exists a neural network
$N_\varepsilon : \mathbb{R}^N \to \mathbb{R}$ such that
\[
\sup_{z \in \pi_N(\mathcal{M})}
\bigl| \Phi_N(z) - N_\varepsilon(z) \bigr| < \varepsilon.
\]

Combining the discretization and approximation steps yields
\[
\sup_{x \in \mathcal{M}}
\bigl| \Phi(x) - N_\varepsilon(\pi_N(x)) \bigr| < \varepsilon,
\]
which proves the claim.
\end{proof}

\subsection{Proof of Theorem~\ref{thm:mc} (Consistency of Monte Carlo Radius Estimation)}

\begin{proof}
Let 
\[
r = \sup_{x\in\mathcal{M}} \|x - x_0\|_\infty
\]
and define the empirical estimator:
\[
\hat{r}_n = \max_{1\le i \le n} \|f(s_i,\theta_i) - x_0\|_\infty.
\]

Because $\mathcal{M}$ is compact, the supremum is achieved at some 
$x^\star \in \mathcal{M}$.  
Assuming the sampling distribution has support dense in 
$\mathcal{S}\times\Theta$, with probability one there exists a subsequence 
$(s_{i_k},\theta_{i_k})$ such that
\[
f(s_{i_k},\theta_{i_k}) \to x^\star.
\]

Thus:
\[
\|f(s_{i_k},\theta_{i_k}) - x_0\|_\infty \to \|x^\star - x_0\|_\infty = r.
\]

Since $\hat{r}_n$ is the running maximum, monotone and bounded above by $r$,
it converges almost surely to $r$.
\end{proof}

\subsection{Proof of Proposition~\ref{prop:identification} (Identification as Distance Minimization)}

\begin{proof}
For singleton sets $\{x\}$, the Hausdorff distance reduces to:
\[
d_H(\{x\},\mathcal{M}) = \inf_{y\in\mathcal{M}} \|x-y\|_\infty.
\]

Thus $d_H(\{x\},\mathcal{M}) < \varepsilon$ is equivalent to the existence of 
some $y \in \mathcal{M}$ such that:
\[
\|x - y\|_\infty < \varepsilon,
\]
which is precisely the minimum-distance decision rule in supervised or 
unsupervised classification in Banach spaces.

Hence the perceptual decision reduces to verification of proximity to the 
compact perceptual manifold $\mathcal{M}$.
\end{proof}

\section{Empirical Validation on Perceptual Benchmarks}
To test the universality of the Compact Perceptual Manifold (CPM), we extended our analysis to visual and auditory domains using the MNIST and Spoken MNIST datasets.
\subsection{Geometric Saturation of Digit Manifolds}
\label{sec:mnist_validation}
To demonstrate that the Compact Perceptual Manifold (CPM) framework extends beyond purely physical or physiological signals into complex cognitive representations, we applied our topological analysis to visual and auditory perception using the MNIST and Speech MNIST datasets.

We evaluated the empirical Hausdorff radius for all ten digit classes. If the perceptual manifold of a digit is compact, its empirical radius must saturate rapidly, bounding the deterministic variability of the class.

\begin{figure}[H]
    \centering
    \includegraphics[width=\textwidth]{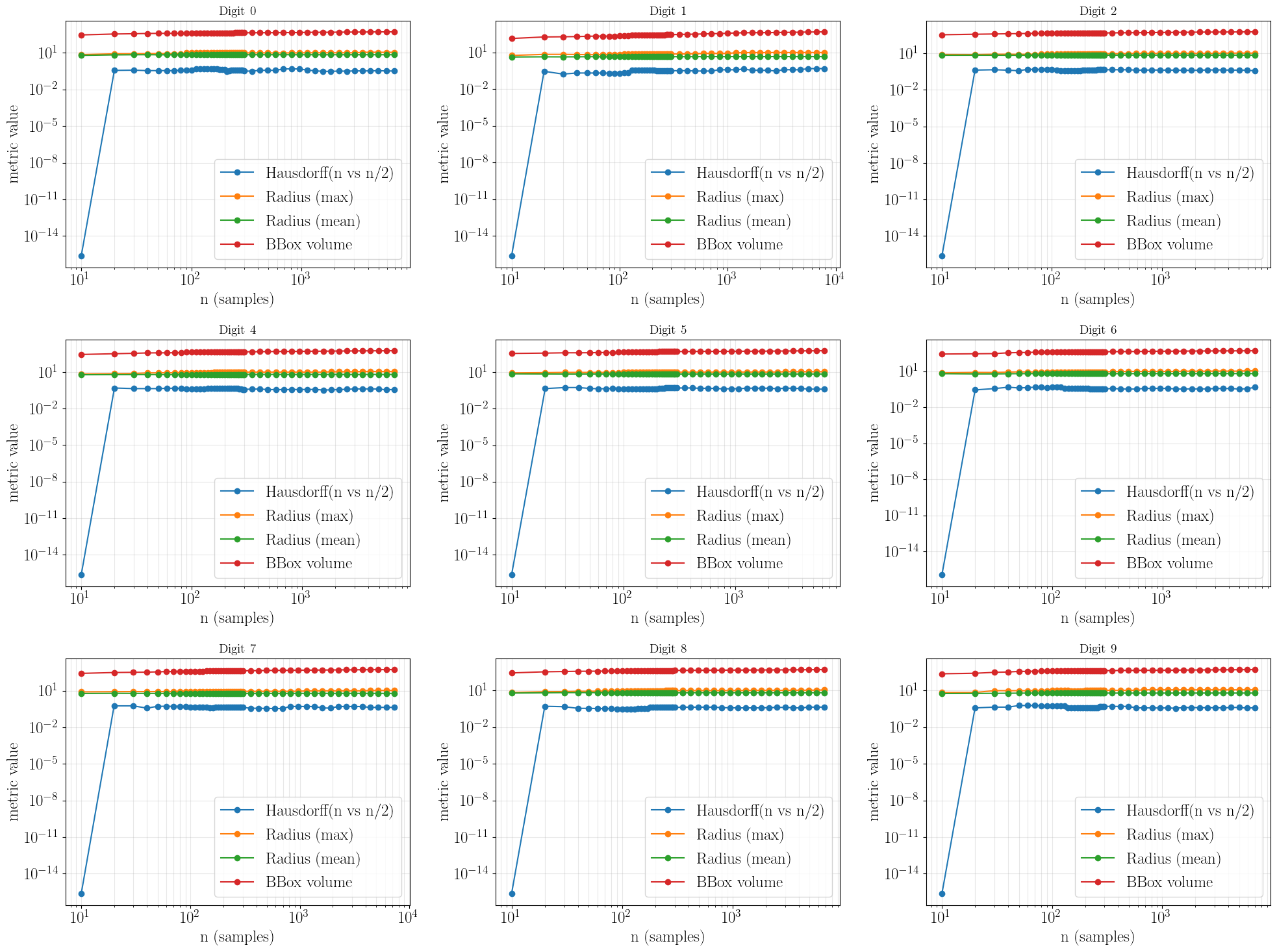} 
    \caption{\small Empirical Hausdorff radius saturation on a particular digit (3) with a minimal number of samples ($n \approx 20$).}
    \label{fig:mnist_10_saturation}
\end{figure}
The rapid saturation observed in Figure \ref{fig:mnist_10_saturation} precisely mirrors the behavior of electromechanical and electrochemical systems, confirming that visual concepts are governed by the same functional-topological invariants.

\begin{figure}[H]
    \centering
    \includegraphics[width=0.95\textwidth]{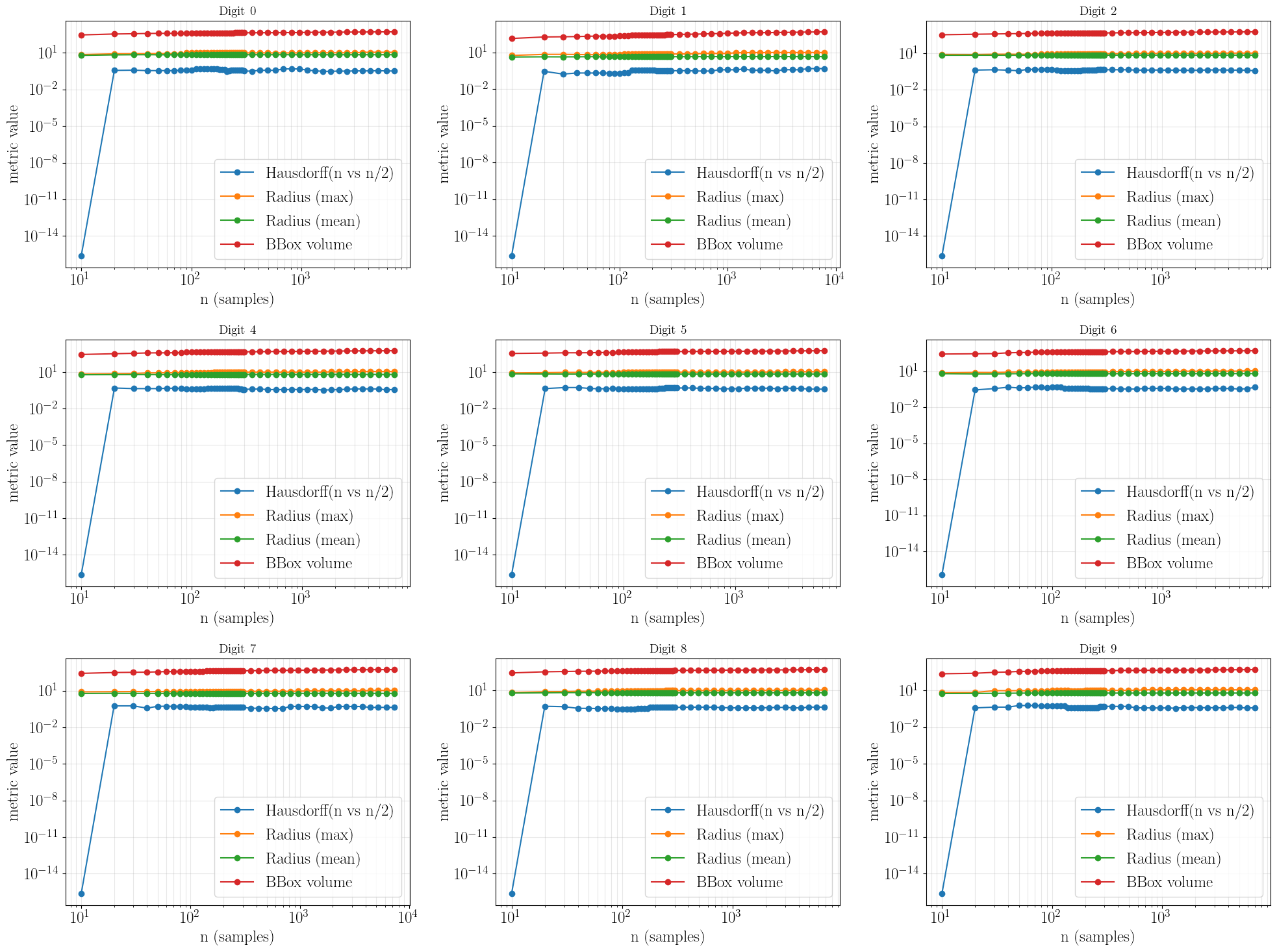} 
    \caption{Empirical Hausdorff radius saturation across all the rest of MNIST digit manifolds. Notice the consistent, rapid asymptotic behavior across all classes, indicating that the structural completeness of the perceptual manifold is achieved with a minimal number of samples ($n \approx 20$).}
    \label{fig:mnist_all_digits_saturation}
\end{figure}

The saturation sample size remains remarkably consistent across the rest of digits,
with $n_{\text{sat}}$ typically between 20 and 30 samples as shown in figure in Figure \ref{fig:mnist_all_digits_saturation}. This indicates
that the perceptual manifolds of handwritten digits exhibit bounded
deterministic variability despite large stylistic differences between
writers.

\section{Multi-site validation on NSRDB solar manifolds}
\label{app:nsrdb_multisite}

To verify that the observed rapid saturation of the solar manifold is
not an artifact of a single location, we repeated the experiment across
multiple NSRDB stations. For each station, we downloaded hourly
\emph{solar-only} NSRDB files from the public NOAA/NCEI archive and
constructed a daily irradiance manifold using exactly the same pipeline
as in the main text: after removing invalid entries and resolving
timestamps of the form \texttt{24:00} as \texttt{00:00} on the following
day, we grouped the global horizontal irradiance (GHI) time series by
calendar day, reindexed each day to a fixed 24-hour local-time grid,
and normalized each daily curve by its maximum value. Each day was thus
represented as a vector in $\mathbb{R}^{24}$, and the collection of such
vectors defined the station-specific solar manifold.

\begin{figure}[ht]
  \centering
  \includegraphics[width=0.8\textwidth]{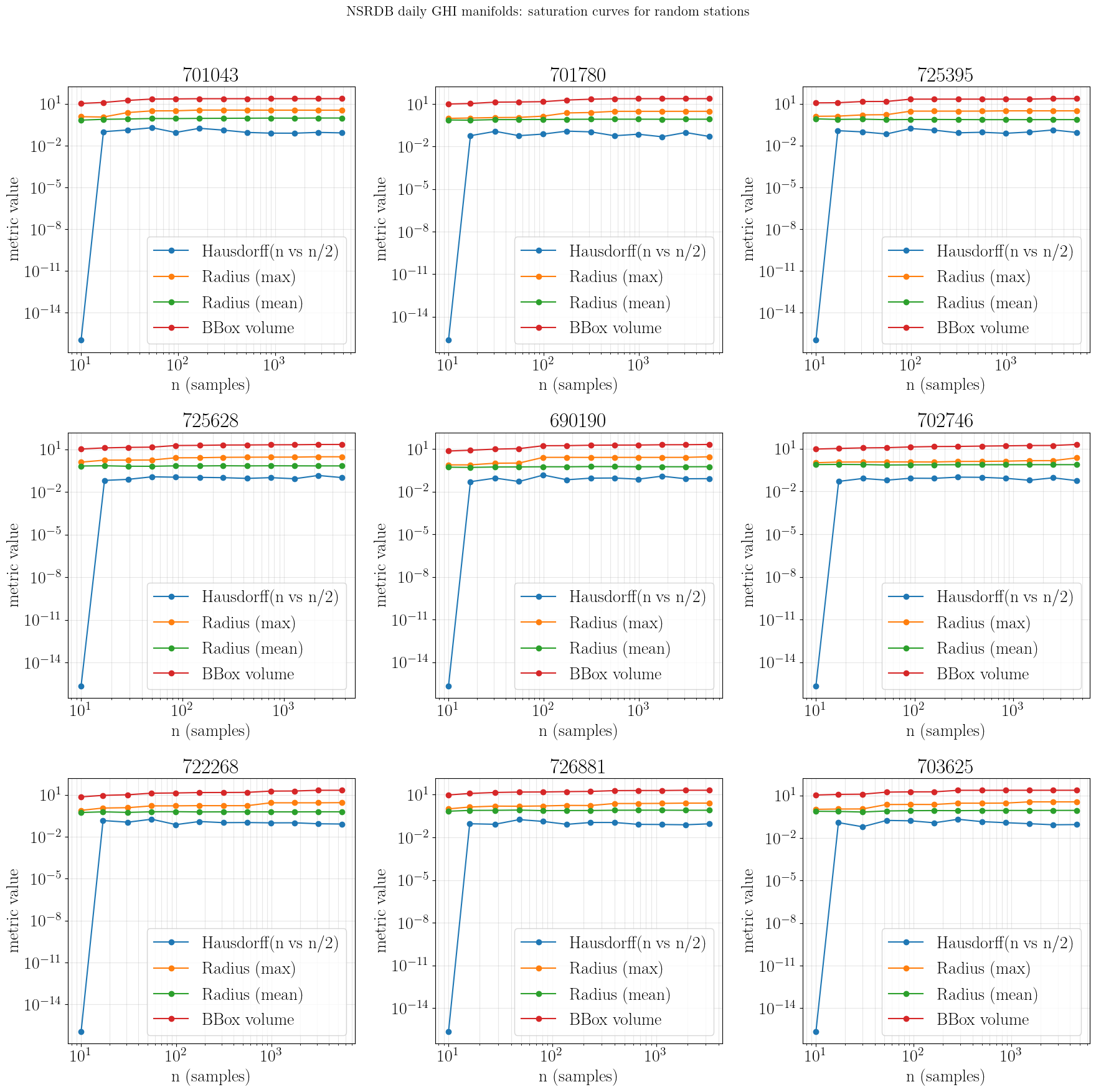}
  \caption{\small Geometric saturation curves for multiple NSRDB solar stations. 
  Each panel represents a different geographical location, showing the 
  consistent emergence of a stable plateau in all geometric metrics.}
  \label{fig:nsrdb_manifold_saturation_grid} 
\end{figure}

To test whether the geometry saturates generically across sites, we
selected 80 stations at random from the downloaded archive and, for each
station, computed the same saturation curves used throughout the paper:
the internal Hausdorff discrepancy between nested samples, the maximum
radius, the mean radius, and the bounding-box volume as functions of
sample size $n$. Figure~\ref{fig:nsrdb_manifold_saturation_grid} shows 
curves for 9 randomly selected stations.  In every displayed case, 
the same qualitative pattern is observed: a rapid initial increase followed
by an extended plateau, indicating that additional daily profiles remain
well inside the previously established envelope. 

To summarize this behavior quantitatively across a larger random sample,
we defined an \emph{operational saturation scale} $n_{\mathrm{sat}}$ for
each station using the mean-radius curve. Specifically, $n_{\mathrm{sat}}$
was taken as the first sample size at which the relative change in mean
radius between successive steps fell below 5\% for two consecutive
steps. Under this criterion, the median saturation scale across the
sample was 22 days, with 68.75\% of stations saturating by 30 days and
81.25\% by 40 days. These values support the claim that, for fixed-site
solar irradiance, only a few tens of daily profiles are typically
required to resolve the geometry of the perceptual manifold.

\begin{table}[htbp]
\centering

\begin{minipage}[t]{0.44\textwidth}
\centering
\small
\vspace{25pt}
\begin{tabular}{lc}
\toprule
\textbf{Statistic} & \textbf{Value} \\
\midrule
Number of stations & 80 \\
Mean $n_{\mathrm{sat}}$ & 27.83 days \\ 
Std. dev. & 22.73 days \\
Median $n_{\mathrm{sat}}$ & 22 days \\
25th percentile & 10 days \\
75th percentile & 36 days \\
Minimum & 10 days \\
Maximum & 129 days \\
Fraction with $n_{\mathrm{sat}} \leq 30$ & 68.75\% \\
Fraction with $n_{\mathrm{sat}} \leq 40$ & 81.25\% \\
\bottomrule
\end{tabular}

\vspace{1.5mm}
\textbf{(a)} Summary statistics for $n_{\mathrm{sat}}$.
\end{minipage}
\hfill
\begin{minipage}[t]{0.50\textwidth}
\centering
\vspace{0pt}
\includegraphics[width=\textwidth,height=0.30\textheight,keepaspectratio]{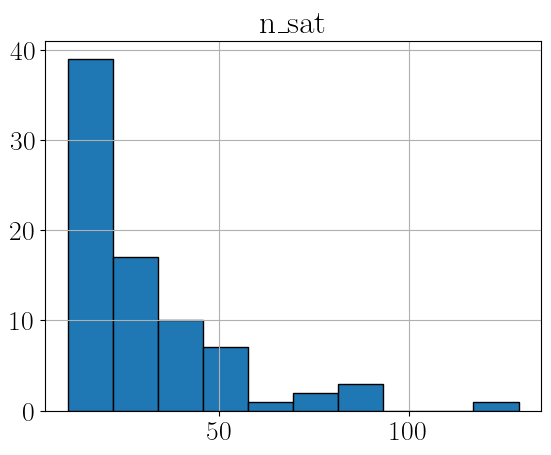}
\label{fig:nsrdb_nsat_distribution}
\vspace{1.5mm}
\textbf{(b)} Histogram of $n_{\mathrm{sat}}$ across stations.
\end{minipage}

\caption{\small Distribution of the operational saturation scale
$n_{\mathrm{sat}}$ across $N=80$ randomly sampled NSRDB stations. The
left panel reports aggregate statistics, while the right panel visualizes
the right-skewed distribution. Most sites saturate within a few tens of
daily profiles, with a visible right tail of larger values.}
\label{tabfig:nsrdb_nsat_summary}
\end{table}

Although the distribution exhibits a visible right tail, its bulk is
concentrated in the regime of a few tens of daily profiles. Thus, the
representative saturation scale for fixed-site daily irradiance remains
small even in the presence of occasional late-saturating stations.

\section{Multi-site validation on NOAA tidal manifolds}
\label{app:noaa_multisite}

\begin{figure}[ht]
  \centering
  \includegraphics[width=0.8\textwidth]{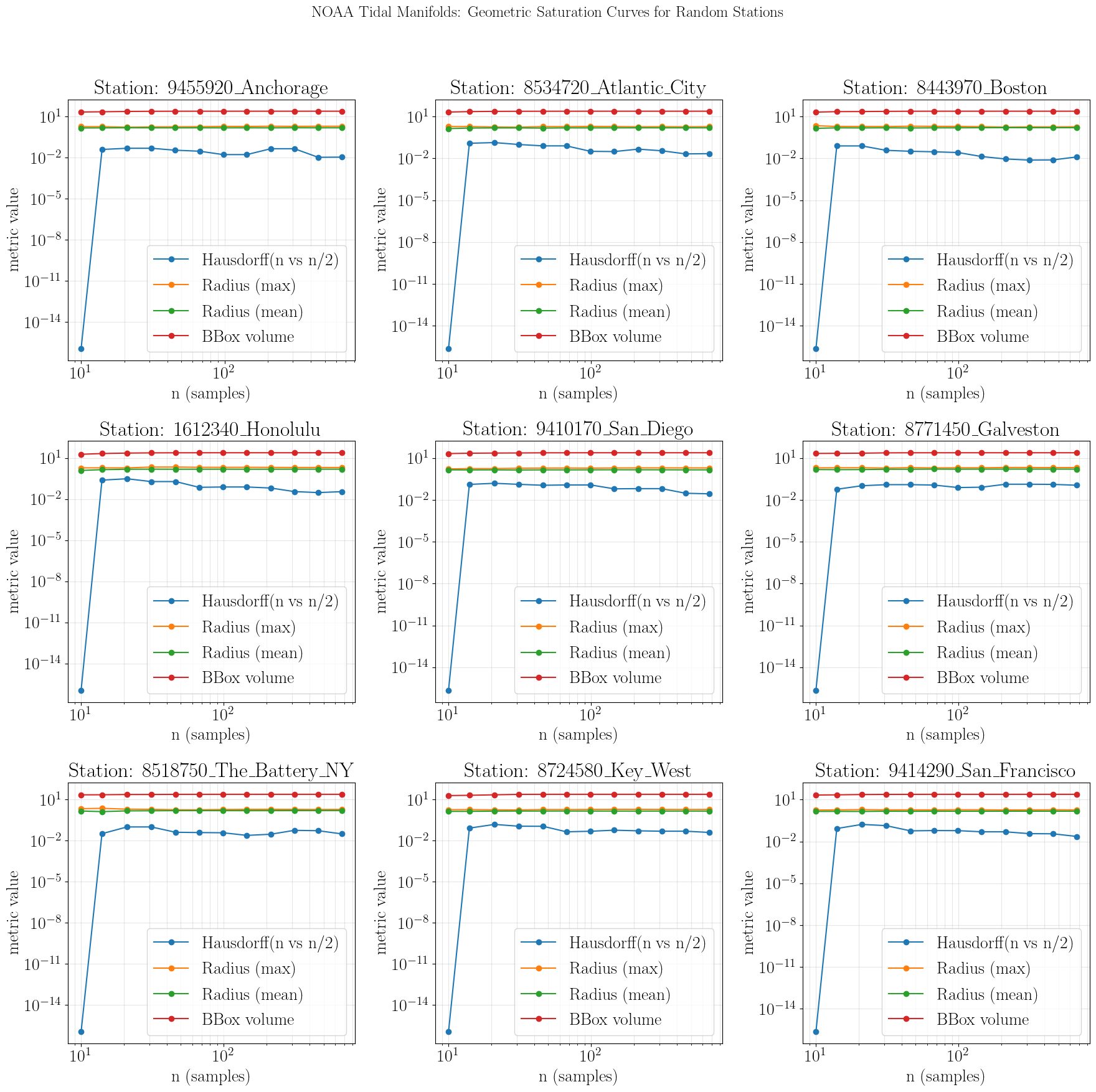}
  \caption{\small Geometric saturation curves for multiple NOAA tidal stations. Each panel corresponds to a different coastal station. The near-instantaneous stabilization of the Hausdorff radius across all sites confirms that the tidal "Blueprint" is exceptionally compact and low-dimensional.}
  \label{fig:tides_multisite_saturation} 
\end{figure}

To verify that the rapid geometric saturation observed in geophysical signals is a general property of tidal dynamics rather than a site-specific artifact, we extended our analysis across multiple coastal environments. We utilized the NOAA Center for Operational Oceanographic Products and Services (CO-OPS) API to retrieve verified hourly water-level data from a geographically diverse set of stations, including Pacific (Honolulu, San Francisco), Atlantic (Boston, Charleston), and Gulf Coast (Galveston) locations. 
For each station, we constructed a daily tidal manifold using a pipeline analogous to the one described in the main text: the continuous sea-level time series was segmented into 24-hour calendar days, reindexed to a fixed hourly grid, and processed to resolve any missing observations via linear interpolation. To isolate the topological structure of the tidal harmonics from site-specific tidal ranges and mean sea-level offsets, each daily curve was normalized using min-max scaling, mapping the values to the unit interval $[0, 1]$. Each day was thus represented as a vector in $\mathbb{R}^{24}$, and the collection of these vectors defined the station-specific tidal manifold.
To test whether the geometry saturates generically across different harmonic regimes (e.g., diurnal, semi-diurnal, and mixed tides), we computed the same saturation curves used throughout this work: the internal Hausdorff discrepancy, the maximum radius, the mean radius, and the bounding-box volume as functions of sample size $n$. Figure~\ref{fig:tides_multisite_saturation} shows these curves for the 9 selected stations. In every case, we observe an exceptionally abrupt collapse of the geometric metrics, with the mean radius reaching a stable plateau significantly faster than in the solar or physiological domains. This suggests that the deterministic astronomical forcing is so dominant that the manifold's envelope is exhausted almost as soon as a single spring-neap cycle is observed.

To summarize this behavior quantitatively, we applied the operational saturation scale $n_{\mathrm{sat}}$ (defined as the first sample size where the relative change in mean radius stays below 5\% for two consecutive steps). The aggregate statistics, reported in Table~\ref{tabfig:tides_nsat_summary}, reveal a remarkably efficient characterization of the manifold: the median saturation scale is only 10 days, meaning that less than two weeks of data are required to resolve the full geometry of the local tidal system.

\begin{table}[htbp]
\centering
\begin{minipage}[t]{0.44\textwidth}
\centering
\small
\vspace{25pt}
\begin{tabular}{lc}
\toprule
\textbf{Statistic} & \textbf{Value} \\
\midrule
Number of stations & 9 \\
Mean $n_{\mathrm{sat}}$ & 16.00 days \\ 
Std. dev. & 7.75 days \\
Median $n_{\mathrm{sat}}$ & 10.00 days \\
25th percentile & 10.00 days \\
75th percentile & 21.00 days \\
Minimum & 10.00 days \\
Maximum & 31.00 days \\
Fraction with $n_{\mathrm{sat}} \leq 15$ & 55.56\% \\
Fraction with $n_{\mathrm{sat}} \leq 31$ & 100.0\% \\
\bottomrule
\end{tabular}

\vspace{1.5mm}
\textbf{(a)} Summary statistics for $n_{\mathrm{sat}}$.
\end{minipage}
\hfill
\begin{minipage}[t]{0.50\textwidth}
\centering
\vspace{0pt}
\includegraphics[width=\textwidth,height=0.30\textheight,keepaspectratio]{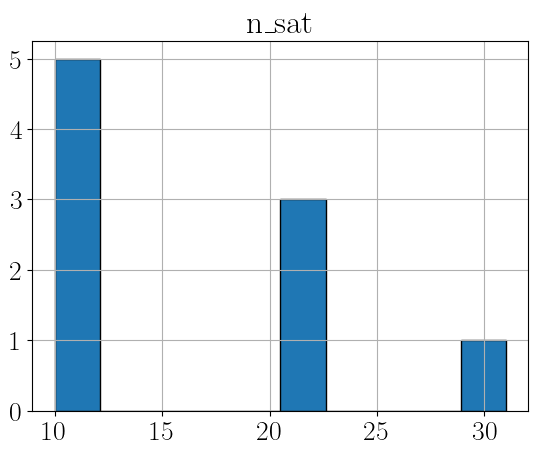}

\vspace{1.5mm}
\textbf{(b)} Histogram of $n_{\mathrm{sat}}$ across tidal stations.
\end{minipage}

\caption{\small Distribution of the operational saturation scale $n_{\mathrm{sat}}$ across $N=9$ NOAA tidal stations. The distribution is notably shifted to the left compared to other domains, with all stations reaching saturation within a single lunar month ($n \leq 31$).}
\label{tabfig:tides_nsat_summary}
\end{table}

The contrast between these results and the solar manifolds is instructive. While solar irradiance is subject to stochastic atmospheric interference (clouds) that introduces late-stage variability, tidal levels are governed by the high-precision determinism of gravitational mechanics. Consequently, the tidal manifold "fills" almost twice as fast as the solar one. These findings reinforce the core thesis of this work: the more a signal is governed by fundamental physical constraints, the more compact its "Blueprint" becomes, allowing for near-instantaneous geometric identification.
\section{Synthetic Generator for the Electromechanical Domain}
\label{app:gen_electromechanical}

This appendix details the deterministic generator used to construct synthetic 
electromechanical traces.  
Its purpose is not to reproduce the detailed physics of any specific machine, 
but to create a compact, bounded-variability functional manifold with the 
characteristic morphology observed in real electromechanical actuation: an 
initial idle regime, a fast high-amplitude transient, a quasi-stationary 
plateau, and a terminal decay.  
The generator produces only the AC waveform.  
The RMS envelopes used in the experiments are computed from these synthetic AC 
signals using exactly the same sliding-window RMS operator applied to the 
real electromechanical data.  
This guarantees that the synthetic and real RMS traces are directly comparable 
and that no domain-specific preprocessing differences bias the geometric 
analysis.

\subsection{Time segmentation}

Let the actuation interval $[0,T]$ be partitioned into the ordered times
\[
0 < t_{\mathrm{idle}} < t_{\mathrm{peak}} < t_{\mathrm{end}} <
t_{\mathrm{cut}} < t_{\mathrm{step}} < t_{\mathrm{ramp}} < T,
\]
defined from a compact parameter vector
\[
\theta = (A_{\mathrm{idle}}, A_{\mathrm{peak}}, A_{\mathrm{plateau}}, 
          \Delta t_{\mathrm{idle}}, \Delta t_{\mathrm{rise}},
          \Delta t_{\mathrm{decay}}, t_{\mathrm{cut}},
          \Delta t_{\mathrm{step}}, \Delta t_{\mathrm{ramp}}) \in \Theta.
\]
The boundaries are
\[
\begin{aligned}
t_{\mathrm{idle}} &= \Delta t_{\mathrm{idle}}, \\
t_{\mathrm{peak}} &= t_{\mathrm{idle}} + \Delta t_{\mathrm{rise}},\\
t_{\mathrm{end}}  &= t_{\mathrm{peak}} + \Delta t_{\mathrm{decay}},\\
t_{\mathrm{step}} &= t_{\mathrm{cut}} + \Delta t_{\mathrm{step}},\\
t_{\mathrm{ramp}} &= t_{\mathrm{step}} + \Delta t_{\mathrm{ramp}}.
\end{aligned}
\]
All parameters vary in bounded ranges, so $\Theta$ is compact.

\subsection{Deterministic envelope}

A piecewise envelope $e_\theta(t)$ is constructed as follows.

\paragraph{Idle and micro-bump.}
A baseline amplitude $A_{\mathrm{idle}}$ is maintained on $[0,t_{\mathrm{idle}})$, 
with a small pre-transient bump of the form
\[
b(t) = A_{\mathrm{bump}}
       \sin \bigl( \varphi(t) \bigr), \qquad 
\varphi(t) \in [0,\pi/2],
\]
applied over a short interval ending at $t_{\mathrm{idle}}$.

\paragraph{Fast rise.}
On $[t_{\mathrm{idle}}, t_{\mathrm{peak}})$ the envelope increases linearly from 
the end of the bump to the peak amplitude $A_{\mathrm{peak}}$:
\[
e_\theta(t)
  = \text{lin}\bigl(t;\, A_{\mathrm{bump,end}},\, A_{\mathrm{peak}}\bigr).
\]

\paragraph{Shoulder and drop.}
A short plateau around $A_{\mathrm{peak}}$ is followed by a linear decrease to
\[
A_{\mathrm{drop}} = \alpha\,A_{\mathrm{peak}},\qquad \alpha \in (0,1),
\]
over a compact-duration window.

\paragraph{Decay to plateau.}
Over $[t_{\mathrm{shoulder,end}}, t_{\mathrm{end}})$ the envelope decays 
linearly to the plateau amplitude $A_{\mathrm{plateau}}$.

\paragraph{Fast cutoff and slow decay.}
At $t_{\mathrm{cut}}$ a fast drop occurs to
\[
A_{\mathrm{step}} = \beta A_{\mathrm{plateau}}, \qquad \beta \in (0,1),
\]
followed by a slow exponential decay:
\[
e_\theta(t) = A_{\mathrm{step}} e^{-k(t - t_{\mathrm{step}})}, 
\qquad t_{\mathrm{ramp}} \le t \le T,
\]
with $k>0$ bounded.

\subsection{AC carrier and observable waveform}

The synthetic observable is the full AC waveform
\[
x_\theta(t)
  = e_\theta(t)\,\sin(2\pi f_{\mathrm{ac}} t + \phi)
    + \eta_\theta(t),
\]
where:
\begin{itemize}
\item $f_{\mathrm{ac}}$ is the AC carrier frequency,
\item $\phi$ is a bounded phase,
\item $\eta_\theta(t)$ is a bounded perturbation term consisting of:
  \[
  \eta_\theta(t)
    = \eta^{\mathrm{white}}_\theta(t)
    + \eta^{\mathrm{mult}}_\theta(t)
    + \eta^{\mathrm{OU}}_\theta(t),
  \]
  combining small white noise, multiplicative amplitude-dependent noise,  
  and a temporally correlated Ornstein–Uhlenbeck component.
\end{itemize}

\subsection{RMS computation}

No RMS envelope is generated analytically.  
Instead, the RMS signal used for geometric analysis is computed directly from 
$x_\theta(t)$ using the same sliding-window RMS operator applied to the 
real electromechanical data.  
This preserves complete methodological consistency between real and synthetic 
signals and prevents introducing artifacts from manual RMS design.

\subsection{Compactness}

Since $\Theta$ is compact and $\theta \mapsto x_\theta(\cdot)$ is continuous in 
the supremum norm, the synthetic manifold
\[
\mathcal{M}_{\mathrm{elec}}
 = \{ x_\theta : \theta \in \Theta \}
\]
is compact in $C^0([0,T])$ and therefore admits a finite Hausdorff radius.

\subsection*{Implementation} 

A reference implementation of this generator, including the AC waveform, noise 
components, and RMS computation pipeline, are available in the public repository
associated with this work.

\section{Synthetic Generator for the Electrochemical Domain}
\label{app:gen_battery}
\subsection{Electrochemical discharge structure}

Battery discharge under fixed load conditions produces voltage curves with
characteristic properties:
\begin{itemize}
    \item monotonic or quasi-monotonic decay,
    \item smooth curvature with a knee region,
    \item bounded voltage range determined by chemistry and operating limits,
    \item absence of high-frequency transients.
\end{itemize}

These properties reflect deterministic electrochemical constraints governed by
reaction kinetics, internal resistance, and diffusion processes.

\subsection{Parameterization}

Each discharge curve is parameterized by a vector
\[
\theta = (V_0, V_{\min}, \alpha, \beta, t_k, \gamma, \varepsilon) \in \Theta,
\]
where:
\begin{itemize}
    \item $V_0$ is the initial voltage,
    \item $V_{\min}$ is the cutoff voltage,
    \item $\alpha$ controls early-stage decay,
    \item $\beta$ controls mid-stage curvature,
    \item $t_k$ defines the knee location,
    \item $\gamma$ controls terminal tapering,
    \item $\varepsilon$ bounds smooth perturbations.
\end{itemize}

All parameters vary within bounded intervals, so $\Theta$ is compact.

\subsection{Deterministic discharge model}

The nominal discharge envelope is defined as
\[
v_\theta(t) =
V_{\min}
+ (V_0 - V_{\min})
\exp\!\left(
- \alpha t
- \beta \max(0, t - t_k)^2
\right),
\qquad t \in [0,T].
\]

This formulation captures:
\begin{itemize}
    \item an initial exponential decay,
    \item a curvature increase around the knee,
    \item a smooth approach to the cutoff voltage.
\end{itemize}

\subsection{Smooth bounded perturbations}

To account for structured variability while preserving continuity, we add a
bounded smooth perturbation:
\[
\eta_\theta(t)
= \varepsilon \sum_{j=1}^{J} c_j(\theta)\,\psi_j(t),
\]
where:
\begin{itemize}
    \item $\psi_j$ are fixed smooth basis functions on $[0,T]$,
    \item $c_j(\theta)$ are bounded continuous coefficients,
    \item $J$ is finite.
\end{itemize}

The final observable voltage curve is
\[
x_\theta(t) = v_\theta(t) + \eta_\theta(t).  
\]

\subsection{Compactness of the electrochemical manifold}

The synthetic electrochemical manifold is defined as
\[
\mathcal{M}_{\mathrm{bat}}
= \{\, x_\theta(\cdot) : \theta \in \Theta \,\}
\subset C^0([0,T]).
\]

Since $\Theta$ is compact and the map
$\theta \mapsto x_\theta(\cdot)$ is continuous in the supremum norm, the image
$\mathcal{M}_{\mathrm{bat}}$ is compact by the standard image-of-compact-is-compact
argument. Consequently, the electrochemical perceptual manifold admits a finite
Hausdorff radius.

\subsection{Status of synthetic evaluation}

The electrochemical domain is empirically validated in this work using real
NASA battery discharge curves, as reported in Section~6.2. The synthetic
electrochemical generator described above is included only as a theoretical
reference model illustrating how bounded discharge families may arise from a
compact parameterization.
We do not use this synthetic generator in the reported battery experiments.
Unlike the point-machine and ECG cases, where synthetic generators are used to
compare real and simulated manifold saturation, the battery results in this work
are based exclusively on real discharge data. A more detailed physics-aware
battery simulator is left for future work.


\section{Synthetic Generators for the ECG Domain}
\label{app:gen_ecg}

This appendix details the synthetic generators used to construct compact
functional manifolds representative of electrocardiogram (ECG) signals.
The goal is not to reproduce the full electrophysiological complexity of
cardiac dynamics, but to generate bounded, continuous families of waveforms
with stereotyped heartbeat morphology suitable for geometric analysis.

Two generators of increasing abstraction are considered:
(i) a dynamical ECG generator based on the McSharry model, and
(ii) a purely morphological Gaussian generator.
Both produce continuous signals on a fixed time interval and induce compact
perceptual manifolds in $C^0([0,T])$.

\subsection{Common preprocessing and representation}

All synthetic ECG signals are generated on a fixed interval $[0,T]$ centered
around the R-peak. Signals are resampled onto a uniform temporal grid and
normalized in amplitude using the same preprocessing pipeline applied to real
ECG data. This guarantees that synthetic and real signals are directly
comparable under the same metric geometry.

\subsection{McSharry dynamical ECG generator}

The McSharry generator~\cite{mcsharry2003dynamical} models ECG signals using a
low-dimensional nonlinear dynamical system designed to produce a stereotyped
P--QRS--T morphology. While simplified, this model captures the gross temporal
structure of cardiac cycles.

\subsubsection{Parameterization}

Each realization is determined by a parameter vector
\[
\theta = (A_P, A_Q, A_R, A_S, A_T;\;
          t_P, t_Q, t_R, t_S, t_T;\;
          \sigma_P, \sigma_Q, \sigma_R, \sigma_S, \sigma_T;\;
          \varepsilon)
\in \Theta,
\]
where:
\begin{itemize}
    \item $A_k$ control the amplitudes of the P, Q, R, S, and T components,
    \item $t_k$ define their temporal locations,
    \item $\sigma_k$ control their temporal widths,
    \item $\varepsilon$ bounds smooth perturbations.
\end{itemize}
All parameters vary within bounded intervals, so $\Theta$ is compact.

\subsubsection{Deterministic waveform construction}

The nominal ECG waveform is constructed as a superposition of smooth
components:
\[
f_\theta(t) = \sum_{k \in \{P,Q,R,S,T\}}
A_k \exp\!\left( -\frac{(t - t_k)^2}{2\sigma_k^2} \right).
\]

Additional smooth perturbations are added to account for structured variability:
\[
\eta_\theta(t)
= \varepsilon \sum_{j=1}^{J} c_j(\theta)\,\psi_j(t),
\]
where $\psi_j$ are fixed smooth basis functions on $[0,T]$ and $J$ is finite.

The final observable signal is
\[
x_\theta(t) = f_\theta(t) + \eta_\theta(t).
\]

\subsubsection{Compactness}

Since $\Theta$ is compact and the map
$\theta \mapsto x_\theta(\cdot)$ is continuous in the supremum norm, the
synthetic ECG manifold generated by the McSharry model,
\[
\mathcal{M}_{\mathrm{ECG}}^{\mathrm{McS}}
= \{ x_\theta : \theta \in \Theta \},
\]
is compact in $C^0([0,T])$ and admits a finite Hausdorff radius.

\subsection{Gaussian morphological ECG generator}

To decouple geometric properties from physiological modeling assumptions, we
also consider a purely morphological ECG generator based on Gaussian
components. This generator ignores cardiac dynamics entirely and retains only
coarse waveform structure.

\subsubsection{Parameterization}

Each synthetic beat is defined by
\[
\theta = (a_1,a_2,a_3;\; t_1,t_2,t_3;\;
          \sigma_1,\sigma_2,\sigma_3;\; b,\; \varepsilon)
\in \Theta,
\]
where:
\begin{itemize}
    \item $a_k$ control the amplitudes of three dominant excursions
          (e.g.\ P/QRS/T),
    \item $t_k$ define their temporal locations,
    \item $\sigma_k$ control their widths,
    \item $b$ defines a baseline offset,
    \item $\varepsilon$ bounds smooth perturbations.
\end{itemize}
All parameters are bounded, so $\Theta$ is compact.

\subsubsection{Deterministic waveform}

The deterministic component is
\[
f_\theta(t)
= b + \sum_{k=1}^{3}
a_k \exp\!\left( -\frac{(t - t_k)^2}{2\sigma_k^2} \right),
\]
with a smooth perturbation term $\eta_\theta(t)$ defined as in the previous
section. The observable waveform is
\[
x_\theta(t) = f_\theta(t) + \eta_\theta(t).
\]

\subsubsection{Compactness}

As before, compactness of $\Theta$ and continuity of the construction imply
that the Gaussian ECG manifold
\[
\mathcal{M}_{\mathrm{ECG}}^{\mathrm{Gauss}}
= \{ x_\theta : \theta \in \Theta \}
\]
is compact in $C^0([0,T])$ and admits a finite Hausdorff radius.

\subsection{Remarks on realism and geometry}

Neither generator is intended to produce physiologically accurate ECG signals.
The McSharry model captures coarse heartbeat dynamics but omits many biological
details, while the Gaussian generator is purely morphological.

Nevertheless, both generators produce compact functional manifolds whose
geometric saturation properties closely match those observed in real ECG data.
This demonstrates that early saturation and finite perceptual radius are
geometric consequences of bounded deterministic variability, not of detailed
physiological realism.

\subsection*{Implementation}

Reference implementations of both ECG generators, together with preprocessing
and resampling routines identical to those applied to real ECG data, will be
provided in the public repository associated with this work.

\end{document}